\documentclass{article}
\usepackage{placeins}
\usepackage[numbers, compress]{natbib}
 \usepackage{fontawesome5} 
\usepackage[eandd, final]{neurips_2026}

\usepackage[utf8]{inputenc} %
\usepackage[T1]{fontenc}    %
\usepackage{hyperref}       %
\usepackage{url}            %
\usepackage{booktabs}       %
\usepackage{amsfonts}       %
\usepackage{nicefrac}       %
\usepackage{microtype}      %
\usepackage{xcolor}         %
\usepackage{amssymb}
\usepackage{pifont}
\usepackage{multirow}
\usepackage{graphicx}
\usepackage{threeparttable}
\usepackage[table]{xcolor}
\usepackage{booktabs}
\usepackage{multirow}
\usepackage{makecell}
\usepackage{array}
\usepackage{booktabs}
\usepackage{makecell}
\usepackage{wrapfig}
\usepackage{subfigure}
\usepackage{tabularx}
\usepackage{enumitem}
\usepackage{fancyvrb}
\usepackage{fvextra}
\definecolor{avgcolor}{RGB}{255, 245, 220} %
\definecolor{overallcolor}{RGB}{255, 235, 205}
\newcommand{\EMB}{\texttt{EvoMemBench}}
\definecolor{headergray}{RGB}{240,240,240}
\usepackage[most,breakable,listings]{tcolorbox}

\tcbset{
    promptstyle/.style={
        colback=blue!5,    %
        colframe=blue!75,  %
        fonttitle=\bfseries,
        arc=3pt,           %
        outer arc=3pt,
        left=5pt,          %
        right=5pt,
        top=5pt,
        bottom=5pt,
        breakable          %
    }
}

\usepackage{fancyvrb}

\usepackage[most]{tcolorbox}

\DefineVerbatimEnvironment{promptboxapp}{Verbatim}{
    commandchars=\\\{\},
    codes={\catcode`\_=12},
    breaklines=true,
    breaksymbolleft={},
    breaksymbolindentleft=0pt,
    breaksymbolsepleft=0pt
}

\tcolorboxenvironment{promptboxapp}{
    promptstyle
}
\title{\EMB: Benchmarking Agent Memory from a Self-Evolving Perspective}

\author{Yuyao Wang$^{1}$\thanks{Equal contribution}~, Zhongjian Zhang$^{3}$\footnotemark[1]~, Mo Chi$^{1}$\footnotemark[1]~, Kaichi Yu$^{4}$\footnotemark[1]~,  
\and
\textbf{Yuhan Li$^{1}$, Miao Peng$^{1}$, Bing Tong$^{1,2}$, Chen Zhang$^{2}$, Yan Zhou$^{2}$, Jia Li$^{1}$\thanks{Corresponding Author.}}
  \\
$^{1}$Hong Kong University of Science and Technology (Guangzhou), $^{2}$Createlink Technology, \\
$^{3}$Beijing University of Posts and
Telecommunications, $^{4}$Beijing Institute of Technology\\  
\faEnvelope[regular]{} Primary contact: \texttt{jialee@hkust-gz.edu.cn}\\
}

\begin{document}

\maketitle

\begin{abstract}
  Recent benchmarks for Large Language Model (LLM) agents mainly evaluate reasoning, planning, and execution. However, memory is also essential for agents, as it enables them to store, update, and retrieve information over time. This ability remains under-evaluated, largely because existing benchmarks do not provide a systematic way to assess memory mechanisms. In this paper, we study agent memory from a self-evolving perspective and introduce \texttt{EvoMemBench}, a unified benchmark organized along two axes: memory scope (in-episode vs.~cross-episode) and memory content (knowledge-oriented vs.~execution-oriented). We compare 15 representative memory methods with strong long-context baselines under a standardized protocol. Results show that current memory systems are still far from a general solution: long-context baselines remain highly competitive, memory helps most when the current context is insufficient or tasks are difficult, and no single memory form works consistently across all settings. Retrieval-based methods remain strong for knowledge-intensive settings, whereas procedural and long-term memory methods are more effective for execution-oriented tasks when their stored experience matches the task structure. We hope \texttt{EvoMemBench} facilitates future research on more effective memory systems for LLM-based agents. Our dataset and code are available at \url{https://github.com/DSAIL-Memory/EvoMemBench}.

\end{abstract}

\section{Introduction}

Large language models (LLMs) have enabled  increasingly capable agents that reason, plan, and act through interaction with environments and external tools~\citep{du2025deepresearchbenchcomprehensivebenchmark,tan2026gabench}. 
However, their backbone LLMs remain stateless: they can condition on the current input context, but do not natively maintain or update persistent internal state across interactions. 
A common workaround is to append interaction histories, observations, and tool outputs to the prompt~\citep{yao2023react, shinn2023reflexion}, but this approach does not scale as histories grow beyond the finite context window. 
As a result, effective agent memory must go beyond storing past information by continuously updating, organizing, and reusing information as agents interact within and across tasks~\citep{fang2025comprehensivesurveyselfevolvingai, ghareeb2025robinmultiagentautomatingscientific}.

We study this capability as \emph{self-evolving memory}. 
From this perspective, memory should support two complementary forms of adaptation. 
Within an episode, agents must retain and revise evolving knowledge, as well as maintain task-relevant execution state during multi-step interaction. 
Across episodes, agents should accumulate reusable knowledge and distill execution experience that improves future decisions~\citep{fang2026mempexploringagentprocedural}. 
This yields a unified evaluation view along two axes: \emph{in-episode} vs. \emph{cross-episode} evolution, and \emph{knowledge-oriented} vs. \emph{execution-oriented} memory.

Existing memory benchmarks cover only parts of this space. 
Text-centric benchmarks such as LoCoMo~\citep{locomo}, LongMemEval~\citep{wu2024longmemeval}, and MemoryAgentBench~\citep{memoryagentbench} mainly evaluate knowledge retention, retrieval, or revision in conversational or document-like contexts, but do not test whether memory supports action and tool-based execution. 
Recent agentic or lifelong memory benchmarks, including MemoryArena~\citep{he2026memoryarena}, Evo-Memory~\citep{wei2024evomemory}, StuLife~\citep{stulife}, and MemoryBench~\citep{ai2025memorybench}, move closer to an interactive setting, but still do not jointly evaluate in-episode and cross-episode evolution while separating knowledge-oriented and execution-oriented memory demands. 
Moreover, some related benchmarks~\citep{wei2024evomemory,ai2025memorybench} are not open-sourced, limiting reproducible comparison.

To address these gaps, we introduce \EMB, a benchmark for evaluating agent memory from a self-evolving perspective.
\texttt{EvoMemBench}  covers four settings: in-episode knowledge evolution, in-episode execution evolution, cross-episode knowledge evolution, and cross-episode execution evolution. These settings provide a unified testbed for assessing whether memory systems can update and reuse information over time.
Our contributions are summarized as follows:
\begin{itemize}[leftmargin=*, itemsep=2pt, topsep=2pt]
    \item We benchmark agent memory from a self-evolving perspective, organizing memory evaluation along the axes of \textbf{in-episode vs. cross-episode} evolution and \textbf{knowledge vs. execution} demands.

    \item We introduce \EMB, a unified and open benchmark that comprehensively evaluates existing memory systems across four forms of self-evolving memory.

    \item We provide a standardized evaluation protocol and comparative analysis of representative memory methods, revealing when memory helps, when it hurts, and which memory forms best match different agent-memory demands.

\end{itemize}

\begin{figure}[t]
\centerline{\includegraphics[width=\columnwidth]{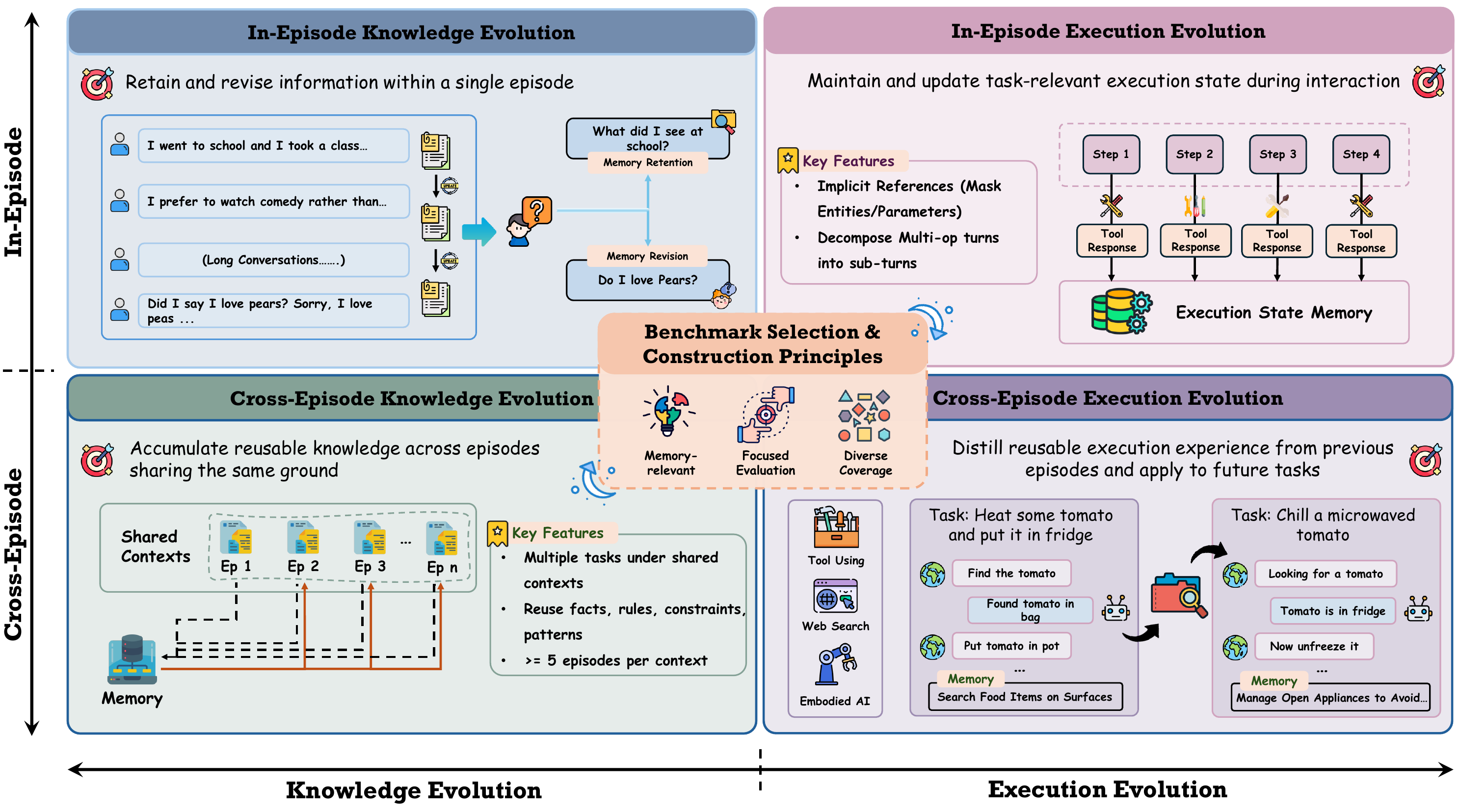}}
\caption{Overview of \EMB.}
\label{fig:timeline}
\label{frame}
\end{figure}

\section{Related Work}

\paragraph{Memory mechanisms for LLM agents.}
Memory is an important component for LLM-based agents, as it allows agents to keep useful information beyond the context window and reuse past experience in later tasks. 
Existing methods mainly differ in what they store and how memory is used. 
Retrieval-based methods store past information in external indexes and retrieve relevant entries during inference~\citep{robertson2009probabilistic,zhang2025qwen3,edge2024local,salama2025meminsightautonomousmemoryaugmentation, du2025memguideintentdrivenmemoryselection,li2026experienceevolvingmultiturntooluseagent,anokhin2025arigraphlearningknowledgegraph}. 
Long-term memory systems further introduce structured storage, updating, and management for persistent facts, preferences, and interaction histories~\citep{chhikara2025mem0, xu2025mem, li2025memos, kang2025memory,nguyen2026byteroveragentnativememoryllmcurated}. 
Procedural memory methods instead abstract trajectories into reusable workflows, skills, or reasoning strategies~\citep{wang2025awm,zheng2025skillweaver,tang2025agent, ouyang2026reasoningbank,han2025legomemmodularproceduralmemory}. 
Recent studies also explore evolving memory architectures~\citep{zhang2026agentic, zhang2025memevolve, zhou2025mementofinetuningllmagents}. 
Together, these methods show that agent memory is moving from passive storage toward reusable experience. 
However, they are usually evaluated on different tasks and protocols, making it hard to compare which memory mechanisms are useful for different self-evolving needs.

\paragraph{Memory benchmarks for LLMs and agents.}
Existing memory benchmarks cover important but partial aspects of agent memory. 
Text-based benchmarks, such as LoCoMo and LongMemEval, evaluate long-term conversational memory through question answering over multi-session histories~\citep{locomo, wu2024longmemeval}, while MemoryAgentBench studies memory agents under incremental multi-turn inputs~\citep{memoryagentbench}. 
More recent benchmarks move toward agentic and continual memory: MemoryBench studies continual learning from simulated user feedback ~\citep{ai2025memorybench}, Evo-Memory evaluates test-time memory evolution over sequential task streams~\citep{wei2024evomemory}, and MemoryArena examines memory-guided action in multi-session environments~\citep{he2026memoryarena}. 
However, as shown in Table~\ref{tab:benchmark_comparison}, existing memory benchmarks do not offer a unified view of how memory evolves across knowledge and execution, within and across episodes.

\begin{table}[t]
\centering
\caption{Comparison between existing memory benchmarks and \EMB. }
\label{tab:benchmark_comparison}
\resizebox{\columnwidth}{!}{
\begin{tabular}{@{}lccccc@{}}
\toprule
\multirow{2}{*}{Benchmark} 
& \multicolumn{2}{c}{In-Episode} 
& \multicolumn{2}{c}{Cross-Episode}
& \multirow{2}{*}{Open-source} \\
\cmidrule(lr){2-3} \cmidrule(lr){4-5}
& Knowledge  Evolution & Execution  Evolution 
& Knowledge  Evolution & Execution  Evolution 
& \\
\midrule
LoCoMo \cite{locomo}          & \textcolor{green!50!black}{\ding{51}} & \textcolor{red}{\ding{55}} & \textcolor{red}{\ding{55}} & \textcolor{red}{\ding{55}} & \textcolor{green!50!black}{\ding{51}} \\
LongMemEval \cite{wu2024longmemeval}     & \textcolor{green!50!black}{\ding{51}} & \textcolor{red}{\ding{55}} & \textcolor{red}{\ding{55}} & \textcolor{red}{\ding{55}} & \textcolor{green!50!black}{\ding{51}} \\
MemoryAgentBench \cite{memoryagentbench}  & \textcolor{green!50!black}{\ding{51}} & \textcolor{red}{\ding{55}} & \textcolor{red}{\ding{55}} & \textcolor{red}{\ding{55}} & \textcolor{green!50!black}{\ding{51}} \\
StuLife \cite{stulife}          & \textcolor{red}{\ding{55}} & \textcolor{green!50!black}{\ding{51}} & \textcolor{green!50!black}{\ding{51}} & \textcolor{green!50!black}{\ding{51}} & \textcolor{green!50!black}{\ding{51}} \\
MemoryBench \cite{ai2025memorybench}      & \textcolor{green!50!black}{\ding{51}} & \textcolor{red}{\ding{55}} & \textcolor{green!50!black}{\ding{51}} & \textcolor{red}{\ding{55}} & \textcolor{red}{\ding{55}} \\
Evo-Memory \cite{wei2024evomemory}       & \textcolor{red}{\ding{55}} & \textcolor{red}{\ding{55}} & \textcolor{green!50!black}{\ding{51}} & \textcolor{green!50!black}{\ding{51}} & \textcolor{red}{\ding{55}} \\
MemoryArena  \cite{he2026memoryarena}     & \textcolor{green!50!black}{\ding{51}} & \textcolor{green!50!black}{\ding{51}} & \textcolor{red}{\ding{55}} & \textcolor{red}{\ding{55}} & \textcolor{green!50!black}{\ding{51}} \\
\midrule
\texttt{EvoMemBench} (ours) 
                  & \textcolor{green!50!black}{\ding{51}} & \textcolor{green!50!black}{\ding{51}} & \textcolor{green!50!black}{\ding{51}} & \textcolor{green!50!black}{\ding{51}} & \textcolor{green!50!black}{\ding{51}} \\
\bottomrule
\end{tabular}}
\end{table}

\section{Problem Formulation}

\subsection{Agents with Memory Mechanisms}
We consider an LLM-based agent that interacts with an environment over a sequence of task episodes, where each episode corresponds to one complete attempt to solve a task instance.
At step \(t\) of episode \(e\), the agent receives an observation \(o_t^{e}\) and has access to the interaction history $h_{t-1}^{e} = (o_1^{e}, a_1^{e}, \ldots, o_{t-1}^{e}, a_{t-1}^{e})$. It then takes an action \(a_t^{e}\) and receives feedback \(y_t^{e}\). For example, in a tool-use task, the action may be a tool call and the feedback is the returned result.
As the episode proceeds, the interaction history grows and may exceed the finite context window of the LLM. 
The agent therefore maintains a memory state \(m_t^{e} \in \mathcal{M}\), which serves as a compact representation of useful information from past interactions. 
At step \(t\), the agent conditions its action on both the current interaction context and the memory state:
$a_t^{e} \sim \pi(\cdot \mid h_{t-1}^{e}, o_t^{e}, m_t^{e}),$
where \(\pi\) denotes the agent policy. After receiving feedback \(y_t^{e}\), the memory state is updated as
$m_{t+1}^{e} = \mathcal{U}(m_t^{e}, o_t^{e}, a_t^{e}, y_t^{e}),$
where \(\mathcal{U}\) denotes the memory update function.  

\subsection{Scope of Memory Evolution: In-Episode and Cross-Episode}

\paragraph{In-episode memory.} Memory is maintained only within a single episode. It summarizes useful information from earlier steps of that episode and is used to support later decisions within the same episode. Thus, in-episode memory is a task-level compact state that retains information from the current episode for subsequent actions.
\paragraph{Cross-episode memory.} Memory is maintained across different episodes. It summarizes useful information from previous task attempts and is used to support decisions in later episodes. Thus, cross-episode memory is an experience-level compact state that retains reusable information from past episodes while avoiding the direct reuse of episode-specific details that are no longer relevant.

\subsection{Content of Memory Evolution: Knowledge and Execution}
\paragraph{Knowledge evolution.}
Knowledge evolution refers to maintaining information that supports later reasoning or answering.
Such information may include facts, constraints, preferences, domain rules, or task patterns.
As new information arrives, memory should preserve valid earlier information and update outdated or contradicted information when later evidence changes what should be remembered.
Thus, knowledge evolution concerns both retaining useful information and revising memory when stored information becomes inaccurate.
\paragraph{Execution evolution.}
Execution evolution refers to the maintenance of information that supports action selection and task progress. Such information can include task progress, intermediate results, tool outputs, unresolved subgoals, previous decisions, action routines, and task-solving procedures. 
As interaction proceeds, memory should help the agent keep track of what has been done, what remains to be done, and which actions are likely to be useful next. 
Thus, execution evolution is concerned with maintaining action-relevant state and reusable execution experience.

\section{Benchmark Construction}

Based on the above forms of memory evolution, we select or reconstruct six datasets to evaluate different aspects of agent memory, as summarized in Table~\ref{tab:dataset}.
Our dataset selection follows three criteria: \emph{memory relevance}, \emph{focused evaluation}, and \emph{coverage}.
First, the tasks should leave room for agent memory to improve performance.
Second, they should involve clear memory dependencies and provide verifiable outcomes.
Third, they should cover diverse domains and task formats to reduce dataset-specific bias.
For clarity, we refer to the six resulting benchmarks as \textsc{InEp-Know}, \textsc{InEp-Exec}, \textsc{CrossEp-Know}, \textsc{CrossEp-Tool}, \textsc{CrossEp-Web}, and \textsc{CrossEp-Emb}.

\subsection{In-Episode Knowledge Evolution}
\textbf{\textsc{InEp-Know}.}
This dataset evaluates whether an agent can retain and revise information within a single episode. 
We build it from the Accurate Retrieval and Selective Forgetting subsets of MemoryAgentBench. 
These subsets are organized in an incremental multi-turn format, where information is provided over time and the agent must update its memory during the episode. 

\subsection{In-Episode Execution Evolution}

\textbf{\textsc{InEp-Exec}.}
This dataset evaluates whether an agent can maintain task-relevant execution state during an ongoing tool-use interaction.
It is built from BFCL-Multiturn-LongContext~\citep{patil2025bfcl}, which contains multi-turn tool-use tasks with noisy context.
We make the dataset more memory-oriented through two reconstruction steps.
First, we replace explicit mentions with implicit references when they can be inferred from previous turns, forcing the agent to recover them from memory.
Second, we split turns with multiple dependent ground-truth actions into simpler sub-turns, so that later sub-turns can refer to earlier objects or states.
The reconstruction is assisted by GPT-5-mini with our designed prompts, and the correctness of the reconstructed samples is manually verified.
Detailed prompts and validation process are provided in Appendix~\ref{app:implementation}.

\subsection{Cross-Episode Knowledge Evolution}
\textbf{\textsc{CrossEp-Know}.}
This dataset evaluates whether an agent can accumulate reusable knowledge across episodes with a shared background.
We build it from CL-Bench~\citep{dou2026cl}, where each context provides new knowledge for multiple context-dependent tasks.
\textsc{CrossEp-Know} covers four knowledge forms: domain facts, rule systems, procedures, and empirical patterns.
Because the required knowledge is provided in the context, the dataset reduces reliance on pre-trained knowledge and focuses on memory accumulation. 
We group CL-Bench samples by shared context and keep only contexts with at least five samples, yielding 120 contexts and 884 samples. We further split these contexts into Easy, Medium, and Hard subsets by tertiles according to the DeepSeek-V3.2 baseline score.
For each context, episodes are processed sequentially: memory is updated after each episode and used in later episodes from the same context. 

\subsection{Cross-Episode Execution Evolution}
Cross-episode execution evolution evaluates whether an agent can learn reusable execution experience from previous episodes and apply it to future tasks. We instantiate this setting with three datasets covering tool use, web search, and embodied interaction.
For each dataset, memory is accumulated and evaluated within each sub-dataset, so that the evaluation focuses on domain-specific execution experience rather than mixing heterogeneous environments. 

\textbf{\textsc{CrossEp-Tool}.}
We reconstruct BFCL-Multiturn-Base following the same procedure as \textsc{InEp-Exec}, and evaluate memory across its four categories. This dataset tests whether memory can capture reusable tool-use patterns, API constraints, and parameter-filling procedures.

\textbf{\textsc{CrossEp-Web}.}
We use xbench-DeepSearch \citep{chen2025xbench} and WebWalkerQA \citep{wu2025webwalker} to evaluate memory in web search tasks. Following MemEvolve \citep{zhang2025memevolve}, we use 170 sampled examples from WebWalkerQA.
This dataset tests whether agents can reuse search and verification experience from previous trajectories.

\textbf{\textsc{CrossEp-Emb}.}
We use the ALFWorld \citep{shridhar2021alfworld} test set and evaluate memory across its six task categories.
This dataset tests whether memory can retain action routines and environment-specific information for later embodied tasks.

\begin{table}[t]
\centering
\caption{Dataset statistics in \EMB, where \#Samples denotes evaluation samples.}
\label{tab:dataset}

\scriptsize
\setlength{\tabcolsep}{3pt}
\renewcommand{\arraystretch}{1}

\begin{tabularx}{\columnwidth}{
@{}
>{\raggedright\arraybackslash}p{0.14\columnwidth}
>{\raggedright\arraybackslash}p{0.16\columnwidth}
>{\raggedright\arraybackslash}p{0.28\columnwidth}
>{\raggedright\arraybackslash}X
@{}
}

\toprule

\textbf{Setting} 
& \textbf{Dataset} 
& \textbf{Source} 
& \textbf{Subset / Domain (\#Samples)} \\

\midrule

\raisebox{-0.8em}{\makecell[l]{In-Episode\\Knowledge}}
& \raisebox{-0.5em}{\textsc{InEp-Know}}
& \raisebox{-0.5em}{MemoryAgentBench}
& \raisebox{-0.5em}{Accurate Retrieval (1000); Selective Forgetting (200)} \\

\midrule

\raisebox{-0.7em}{\makecell[l]{In-Episode\\Execution}}
& \raisebox{-0.5em}{\textsc{InEp-Exec}}
& \raisebox{-0.5em}{\mbox{BFCL-MultiTurn-LongContext}}
& Gorilla File System (50); Vehicle Control (50); Trading Bots (50); Travel Booking (50) \\

\midrule

\raisebox{-1.2em}{\makecell[l]{Cross-Episode\\Knowledge}}
& \raisebox{-1.0em}{\textsc{CrossEp-Know}}
& \raisebox{-1.0em}{CL-Bench}
& Domain Knowledge Reasoning (294); Rule System Application (257); Procedural Task Execution (306); Empirical Discovery \& Simulation (27) \\

\midrule

\multirow{4}{*}[-3.0em]{\makecell[l]{Cross-Episode\\Execution}}

& \raisebox{-0.5em}{\textsc{CrossEp-Tool}}
& \raisebox{-0.5em}{BFCL-MultiTurn-Base}
& Gorilla File System (50); Vehicle Control (50); Trading Bots (50); Travel Booking (50) \\

\cmidrule(lr){2-4}

& \multirow{2}{*}[-0.3em]{\raisebox{0em}{\textsc{CrossEp-Web}}}
& \raisebox{-0.25em}{xbench-DeepSearch}
& Deep Search (100) \\

\cmidrule(lr){3-4}

&
& \raisebox{-0.25em}{WebWalkerQA}
& Web QA (170) \\

\cmidrule(lr){2-4}

& \raisebox{-1.0em}{\textsc{CrossEp-Emb}}
& \raisebox{-1.0em}{ALFWorld}
& Examine in Light (19); Pick\&Place (46); Clean\&Place (37); Cool\&Place (28); Heat\&Place (25); Pick Two\&Place (45) \\

\bottomrule

\end{tabularx}
\end{table}

\section{Experiments}

\subsection{Experiment Setup}

\paragraph{Methods.}
We evaluate memory-free and memory-augmented agents. 
For memory-free agents, we use Gemini-3-Flash~\citep{gemini3flash}, GPT-5-mini~\citep{openai2025gpt5}, and DeepSeek-V3.2~\citep{liu2025deepseek}. 
For memory-augmented agents, we use DeepSeek-V3.2 as the unified backbone and compare five categories of memory methods: retrieval-augmented memory (BM25~\citep{robertson2009probabilistic}, Qwen3-Emb-4B~\citep{zhang2025qwen3}, GraphRAG~\citep{edge2024local}), short-term memory (MemAgent~\citep{yu2025memagent}, MemoBrain~\citep{qian2026memobrain}), general long-term memory (Mem0~\citep{chhikara2025mem0}, A-MEM~\citep{xu2025mem}, MemOS~\citep{li2025memos}, MemoryOS~\citep{kang2025memory}), procedural long-term memory (AWM~\citep{wang2025awm}, SkillWeaver~\citep{zheng2025skillweaver}, AgentKB~\citep{tang2025agent}, ACE \citep{zhang2026agentic}, ReasoningBank~\citep{ouyang2026reasoningbank}), and meta-evolution memory (MemEvolve~\citep{zhang2025memevolve}). 
Full implementation details and applicability criteria are provided in Appendix~\ref{subapp:baselines}.

\paragraph{Metrics.}
We evaluate both task performance and efficiency.
For knowledge-oriented tasks (\textsc{InEp-Know} and \textsc{CrossEp-Know}), we report \textit{answer accuracy}, which measures whether the model produces the correct answer.
For execution-oriented tasks (\textsc{InEp-Exec}, \textsc{CrossEp-Tool}, \textsc{CrossEp-Web}, and \textsc{CrossEp-Emb}), we report \textit{success rate}, which measures whether the agent completes the final goal. We also report \textit{token usage} as an efficiency metric.
\textit{Token usage} measures the total number of LLM inference tokens, including both input and output tokens used by the agent and the memory module.
When comparing methods across multiple subsets, we also report average rank, where a lower rank indicates better overall performance.

\paragraph{Implementation.}
For each dataset, we reuse the inference pipeline or agent framework from the source benchmark whenever possible, and only replace the LLM backbone or attach a memory module.
To ensure a unified implementation, each memory method is wrapped with two interfaces: \texttt{utilize} and \texttt{update}.
The \texttt{utilize} interface takes the current query or state as input and returns task-relevant memory, such as retrieved entries, procedural guidance, or a compact working context, which is then used as part of the agent prompt.
The \texttt{update} interface incorporates newly observed information into the memory, and when it is invoked depends on the evaluation setting.
In in-episode settings, memory is initialized at the beginning of each episode, updated online during the episode, and cleared after the episode ends.
Specifically, memory is updated after each progressive information block in \textsc{InEp-Know} and after each interaction turn in \textsc{InEp-Exec}.
For \textsc{InEp-Exec}, we evaluate four context budgets: 16K, 32K, 64K, and 128K.
When the prompt exceeds the budget, we keep the current task input and memory, and truncate the early history.
In cross-episode settings, memory is updated after each episode using the episode record or full trajectory, and is reused in subsequent episodes from the same context or environment; it is reset when moving to a new independent context or environment group.
For the cross-episode execution setting, we further evaluate cross-environment transfer, where memory is first built on a source environment and then kept frozen during evaluation on a target environment.
Unless a method requires its own embedding model, we use \texttt{text-embedding-v4} from DashScope as the default embedding backend.
We set the retrieval top-$k$ to 10 for knowledge-oriented tasks and 3 for execution-oriented tasks.

\subsection{Main Results}

\subsubsection{In-Episode Knowledge Evolution}

\begin{table*}[t]
\centering
\caption{Answer accuracy (\%) and ranking comparison on \textsc{InEp-Know}. Lower ranks indicate better performance. All memory-augmented methods use DeepSeek-V3.2 as the backbone.}
\label{tab:memory_retention_revision_performance}
\setlength{\tabcolsep}{4pt}
\renewcommand{\arraystretch}{1.1}
\resizebox{\textwidth}{!}{
\begin{tabular}{l|l|cccc>{\columncolor{avgcolor}[\tabcolsep][\tabcolsep]}c|cc>{\columncolor{avgcolor}[\tabcolsep][\tabcolsep]}c|>{\columncolor{overallcolor}[\tabcolsep][\tabcolsep]}c}

\toprule
\multirow{2}{*}[-2.2ex]{Category}
& \multirow{2}{*}[-2.2ex]{Method}
& \multicolumn{5}{c|}{Memory Retention}
& \multicolumn{3}{c|}{Memory Revision}
& \multicolumn{1}{c}{} \\

\cmidrule(lr){3-7}
\cmidrule(lr){8-10}

& & \begin{tabular}[c]{@{}c@{}}Event\\QA\end{tabular}
& \begin{tabular}[c]{@{}c@{}}LME\\(S$^*$)\end{tabular}
& \begin{tabular}[c]{@{}c@{}}Ruler\\qa1\end{tabular}
& \begin{tabular}[c]{@{}c@{}}Ruler\\qa2\end{tabular}
& \multicolumn{1}{c|}{\begin{tabular}[c]{@{}c@{}}Avg.\\Rank\end{tabular}}
& \begin{tabular}[c]{@{}c@{}}FC\\MH\end{tabular}
& \begin{tabular}[c]{@{}c@{}}FC\\SH\end{tabular}
& \multicolumn{1}{c|}{\begin{tabular}[c]{@{}c@{}}Avg.\\Rank\end{tabular}}
& \multicolumn{1}{c}{%
  \raisebox{2.0ex}[0pt][0pt]{%
    \begin{tabular}[c]{@{}c@{}}Overall\\Rank\end{tabular}
  }
} \\
\midrule

\multirow{3}{*}{Long-Context LLM}
& Gemini-3-Flash
& \textbf{98.20} & \textbf{83.00} & \textbf{94.00} & \textbf{77.00} & \textbf{1.00}
& \textbf{58.00} & \textbf{96.00} & \textbf{1.00} & \textbf{1.00} \\
& GPT-5-mini
& 80.60 & 62.33 & 85.00 & 73.00 & 3.50
& 24.00 & 77.00 & 2.00 & 3.00 \\
& DeepSeek-V3.2
& 83.00 & 32.33 & 64.00 & 42.00 & 6.75
& 10.00 & 65.00 & 3.00 & 5.50 \\

\specialrule{0.4pt}{1pt}{2pt}
\specialrule{0.4pt}{0pt}{3pt}

\multirow{3}{*}{\begin{tabular}[c]{@{}l@{}}Retrieval-Augmented\\Memory\end{tabular}}
& BM25
& 88.80 & 50.33 & \underline{73.00} & \underline{55.00} & \textbf{4.25}
& \textbf{7.00} & \textbf{54.00} & \textbf{4.00} & \textbf{4.17} \\
& Qwen3-Emb-4B
& 83.60 & 21.33 & 38.00 & 38.00 & 8.50
& 4.00 & 30.00 & 7.50 & 8.17 \\
& GraphRAG
& 74.20 & 22.00 & 31.00 & 31.00 & 10.50
& 4.00 & 18.00 & 9.50 & 10.17 \\

\midrule
\multirow{2}{*}{Short-Term Memory}
& MemAgent
& 71.80 & 12.67 & 26.00 & 39.00 & 10.75
& 3.00 & 19.00 & 10.00 & 10.50 \\
& MemoBrain
& 77.60 & 10.67 & 30.00 & 32.00 & 11.00
& 1.00 & 20.00 & 11.00 & 11.00 \\

\midrule
\multirow{4}{*}{\begin{tabular}[c]{@{}l@{}}General Long-Term\\Memory\end{tabular}}
& Mem0
& 79.20 & \textbf{52.00} & \textbf{75.00} & \textbf{68.00} & \underline{4.50}
& \textbf{7.00} & \underline{49.00} & \underline{4.50} & \underline{4.50} \\
& A-MEM
& \textbf{91.20} & \underline{51.00} & 60.00 & 46.00 & \textbf{4.25}
& \underline{5.00} & 35.00 & 6.00 & 4.83 \\
& MemOS
& \underline{90.40} & 41.33 & 55.00 & 43.00 & 5.50
& 3.00 & 35.00 & 7.50 & 6.17 \\
& MemoryOS
& 84.80 & 38.67 & 41.00 & 37.00 & 7.50
& 2.00 & 30.00 & 9.50 & 8.17 \\

\bottomrule
\end{tabular}
}
\end{table*}

\begin{wrapfigure}{r}{0.5\textwidth}
\centering
\vspace{-10pt}
\subfigure
{
    \begin{minipage}[b]{1\linewidth}
        \centering
        \includegraphics[width=\textwidth]{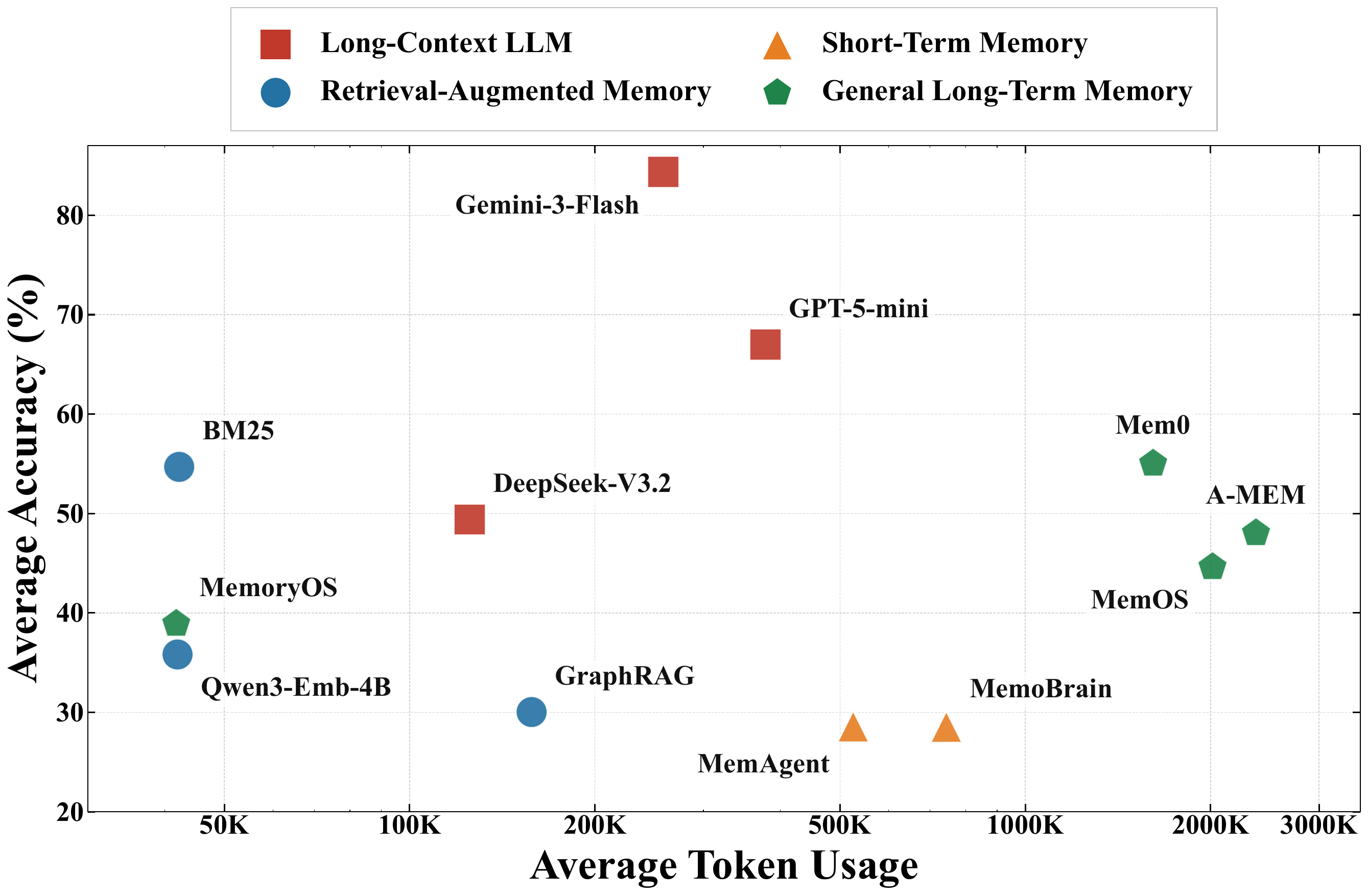}
    \end{minipage}
}
\vspace{-15pt}
\caption{Accuracy and cost on \textsc{InEp-Know}.}
\vspace{-15pt}
\label{fig:perf-cost_trade-off}
\end{wrapfigure}

\paragraph{Finding 1: Strong long-context baselines remain highly competitive.}
As shown in Table~\ref{tab:memory_retention_revision_performance} and Figure~\ref{fig:perf-cost_trade-off}, direct access to the raw context remains a strong baseline.
Gemini-3-Flash achieves the best rank on both retention and revision, suggesting that when the context budget is sufficient, preserving the original evidence may be more reliable than relying on an external memory abstraction.

\paragraph{Finding 2: Existing memory methods are better at retention than revision.}
As shown in Table~\ref{tab:memory_retention_revision_performance}, explicit memory methods show clearer benefits on retention-oriented tasks than on revision-oriented tasks.
BM25, Mem0, and A-MEM improve over the weaker no-memory baseline on retention subsets, but their performance drops sharply on FactConsolidation, especially in the multi-hop setting.
This pattern suggests that the main bottleneck is not merely retrieving relevant evidence.
The harder challenge is deciding which stored information should be updated, suppressed, or replaced when later evidence conflicts with earlier evidence.
Thus, in-episode knowledge evolution requires both effective memory access and robust memory revision.

\subsubsection{In-Episode Execution Evolution}

\begin{table*}[t]
\centering
\caption{Success rate (\%) comparison on \textsc{InEp-Exec} under different context lengths. Best results among memory methods are bolded, and runner-up results are underlined. All memory-augmented methods use DeepSeek-V3.2 as the backbone.}
\label{tab:intra-episode-execution_bfcl_effectiveness_SR}
\setlength{\tabcolsep}{3pt}
\renewcommand{\arraystretch}{1.15}
\resizebox{\textwidth}{!}{
\begin{threeparttable}
\begin{tabular}{l|cccc|cccc|cccc|cccc|>{\columncolor{overallcolor}[\tabcolsep][\tabcolsep]}c>{\columncolor{overallcolor}[\tabcolsep][\tabcolsep]}c>{\columncolor{overallcolor}[\tabcolsep][\tabcolsep]}c>{\columncolor{overallcolor}[\tabcolsep][\tabcolsep]}c}

\toprule
\multirow{2}{*}{Method} 
& \multicolumn{4}{c|}{Gorilla File System}
& \multicolumn{4}{c|}{Vehicle Control}
& \multicolumn{4}{c|}{Trading Bots}
& \multicolumn{4}{c|}{Travel Booking}
& \multicolumn{4}{c}{Overall} \\
\cmidrule(lr){2-5}
\cmidrule(lr){6-9}
\cmidrule(lr){10-13}
\cmidrule(lr){14-17}
\cmidrule(lr){18-21}
& 16K & 32K & 64K & 128K
& 16K & 32K & 64K & 128K
& 16K & 32K & 64K & 128K
& 16K & 32K & 64K & 128K
& \multicolumn{1}{c}{16K}
& \multicolumn{1}{c}{32K}
& \multicolumn{1}{c}{64K}
& \multicolumn{1}{c}{128K} \\
\midrule

\multicolumn{21}{c}{\emph{Long-Context LLM}} \\[2pt]
Gemini-3-Flash 
& \textbf{54.0} & \textbf{50.0} & \textbf{52.0} & \textbf{52.0} 
& \textbf{60.0} & \textbf{70.0} & \textbf{77.6} & \textbf{60.0} 
& \textbf{28.0} & \textbf{80.0} & \textbf{78.0} & \textbf{80.0} 
& \textbf{60.0} & \textbf{62.0} & \textbf{60.0} & \textbf{56.0} 
& \textbf{50.5} & \textbf{65.5} & \textbf{66.9} & \textbf{62.0} \\
GPT-5-mini 
& 34.0 & 34.0 & 40.0 & 34.0 
& 46.0 & 44.0 & 63.7 & 44.0 
& 20.0 & 62.0 & 62.0 & 56.0 
& 40.0 & 30.0 & 32.0 & 36.0 
& 35.0 & 42.5 & 49.4 & 42.5 \\
\rowcolor{headergray}
DeepSeek-V3.2 
& 52.0 & 46.0 & 46.0 & 48.0 
& 38.0 & 40.0 & 73.2 & 44.0 
& 8.0 & 40.0 & 46.0 & 46.0 
& 16.0 & 16.0 & 16.0 & 20.0 
& \cellcolor{overallcolor}28.5 
& \cellcolor{overallcolor}35.5 
& \cellcolor{overallcolor}45.3 
& \cellcolor{overallcolor}39.5 \\

\specialrule{0.4pt}{1pt}{2pt}
\specialrule{0.4pt}{0pt}{3pt}

\multicolumn{21}{c}{\emph{Retrieval-Augmented Memory}} \\[2pt]
BM25 
& 50.0 & \underline{56.0} & 48.0 & 54.0 
& \underline{50.0} & 46.0 & 72.8 & 40.0 
& \underline{20.0} & \underline{54.0} & 48.0 & 42.0 
& 26.0 & 22.0 & 20.0 & 26.0 
& 36.5 & \underline{44.5} & 47.2 & 40.5 \\
Qwen3-Emb-4B 
& 48.0 & \textbf{58.0} & 44.0 & \underline{58.0} 
& 42.0 & \underline{48.0} & 68.7 & 42.0 
& 16.0 & 32.0 & 46.0 & 44.0 
& \underline{32.0} & 20.0 & \underline{26.0} & 24.0 
& 34.5 & 39.5 & 46.2 & 42.0 \\
GraphRAG 
& \underline{56.0} & \underline{56.0} & \underline{54.0} & 54.0 
& 48.0 & 38.0 & \textbf{80.2} & 44.0 
& 12.0 & 24.0 & 46.0 & 42.0 
& 24.0 & \underline{28.0} & \textbf{28.0} & 24.0 
& 35.0 & 36.5 & \underline{52.0} & 41.0 \\

\addlinespace[2pt]
\multicolumn{21}{c}{\emph{Short-Term Memory}} \\[2pt]
MemAgent 
& 48.0 & 48.0 & \underline{54.0} & 46.0 
& 34.0 & 36.0 & \underline{79.8} & 36.0 
& 10.0 & 40.0 & 40.0 & \textbf{52.0} 
& 12.0 & 22.0 & 16.0 & 14.0 
& 26.0 & 36.5 & 47.4 & 37.0 \\
MemoBrain 
& 40.0 & 46.0 & \underline{54.0} & 46.0 
& 40.0 & 32.0 & 38.0 & 42.0 
& 4.0 & 32.0 & 50.0 & 42.0 
& 16.0 & 12.0 & 20.0 & 26.0 
& 25.0 & 30.5 & 40.5 & 39.0 \\

\addlinespace[2pt]
\multicolumn{21}{c}{\emph{General Long-Term Memory}} \\[2pt]
Mem0 
& 54.0 & 44.0 & 46.0 & \textbf{60.0} 
& 46.0 & 44.0 & 70.9 & 38.0 
& 16.0 & 48.0 & 38.0 & 38.0 
& 24.0 & 18.0 & 24.0 & 20.0 
& 35.0 & 38.5 & 44.7 & 39.0 \\
A-MEM 
& 52.0 & 52.0 & 50.0 & 46.0 
& 48.0 & 44.0 & 76.2 & 48.0 
& \textbf{34.0} & 36.0 & 48.0 & 40.0 
& 16.0 & 26.0 & 24.0 & 26.0 
& \underline{37.5} & 39.5 & 49.5 & 40.0 \\
MemOS 
& 52.0 & 52.0 & \underline{54.0} & 50.0 
& 44.0 & 36.0 & 79.0 & 38.0 
& 12.0 & 40.0 & 48.0 & 46.0 
& 22.0 & 12.0 & 14.0 & 16.0 
& 32.5 & 35.0 & 48.8 & 37.5 \\
MemoryOS 
& 48.0 & \textbf{58.0} & 42.0 & 44.0 
& 36.0 & 36.0 & 72.5 & 40.0 
& 10.0 & 38.0 & 42.0 & 20.0 
& 20.0 & 20.0 & 22.0 & 26.0 
& 28.5 & 38.0 & 44.6 & 32.5 \\

\addlinespace[2pt]
\multicolumn{21}{c}{\emph{Procedural Long-Term Memory}} \\[2pt]
AWM 
& 50.0 & 46.0 & 48.0 & 48.0 
& 46.0 & 46.0 & 78.3 & 32.0 
& 10.0 & \textbf{58.0} & \textbf{72.0} & 42.0 
& 20.0 & 22.0 & 14.0 & 18.0 
& 31.5 & 43.0 & \textbf{53.1} & 35.0 \\
SkillWeaver 
& 48.0 & 50.0 & 48.0 & 54.0 
& 44.0 & 42.0 & 73.8 & 46.0 
& 14.0 & 42.0 & 58.0 & 46.0 
& 30.0 & \underline{28.0} & \underline{26.0} & \underline{32.0} 
& 34.0 & 40.5 & 51.5 & \underline{44.5} \\
AgentKB 
& 46.0 & 50.0 & \underline{54.0} & 50.0 
& 8.0 & 40.0 & 77.2 & \textbf{56.0} 
& 18.0 & 26.0 & 56.0 & \underline{50.0} 
& 18.0 & 8.0 & 10.0 & 14.0 
& 22.5 & 31.0 & 49.3 & 42.5 \\
ACE 
& 52.0 & 42.0 & \textbf{56.0} & 54.0 
& 42.0 & 34.0 & 78.4 & 44.0 
& 8.0 & 42.0 & 44.0 & 48.0 
& 26.0 & 22.0 & 24.0 & 20.0 
& 32.0 & 35.0 & 50.6 & 41.5 \\
ReasoningBank 
& \textbf{62.0} & 52.0 & 50.0 & 54.0 
& \textbf{60.0} & \textbf{52.0} & 74.6 & \underline{54.0} 
& 16.0 & 44.0 & 36.0 & 48.0 
& \textbf{34.0} & \textbf{50.0} & \textbf{28.0} & \textbf{36.0} 
& \textbf{43.0} & \textbf{49.5} & 47.1 & \textbf{48.0} \\

\addlinespace[2pt]
\multicolumn{21}{c}{\emph{Meta-Evolution Memory}} \\[2pt]
MemEvolve 
& 52.0 & 48.0 & 46.0 & 54.0 
& 42.0 & 44.0 & 74.8 & 50.0 
& \underline{20.0} & 50.0 & \underline{62.0} & 46.0 
& 30.0 & 22.0 & 20.0 & 22.0 
& 36.0 & 41.0 & 50.7 & 43.0 \\

\bottomrule
\end{tabular}
\end{threeparttable}
}
\end{table*}

\paragraph{Finding 3: Memory helps more when the context budget is constrained.}As reported in Table~\ref{tab:intra-episode-execution_bfcl_effectiveness_SR}, compared with the DeepSeek-V3.2 backbone, the best-performing memory method improves the overall success rate by $+14.5$ points at 16K, $+14.0$ points at 32K, $+7.8$ points at 64K, and $+8.5$ points at 128K.
The largest gains appear at 16K and 32K, where the raw interaction history is more likely to be truncated.
In this regime, memory can compensate for missing history by preserving useful states or execution patterns.
However, as the context budget increases, the marginal benefit of memory becomes smaller and less stable.
At 128K, several memory methods even fall below the no-memory baseline, indicating that appended memory can introduce noise or consume useful context space without adding sufficient information.

\paragraph{Finding 4: Execution tasks benefit from procedural guidance, but compression is risky.}
As reported in Table~\ref{tab:intra-episode-execution_bfcl_effectiveness_SR}, procedural long-term memory methods show the strongest overall pattern.
ReasoningBank achieves the best overall score at 16K, 32K, and 128K, with $43.0$, $49.5$, and $48.0$, respectively, while AWM performs best at 64K with $53.1$.
This suggests that tool-use episodes benefit from reusable execution guidance, such as action routines, failure-aware strategies, or parameter-filling procedures.
At the same time, short-term compression is not always safe.
MemAgent and MemoBrain achieve overall success rates of only $26.0$ and $25.0$ at 16K, both below DeepSeek-V3.2's $28.5$.
At 32K, MemoBrain remains below the baseline ($30.5$ vs. $35.5$), and at 128K, MemAgent also underperforms the baseline ($37.0$ vs. $39.5$).
This suggests that execution-state tracking requires fine-grained details, such as exact tool-call arguments, entity references, and intermediate results.

\subsubsection{Cross-Episode Knowledge Evolution}

\begin{table*}[htbp]
\centering
\caption{Answer accuracy (\%) comparison on \textsc{CrossEp-Know} under different difficulty levels. Best results are bolded, and runner-up results among memory methods are underlined. All memory-augmented methods use DeepSeek-V3.2 as the backbone.}
\label{tab:task_type_difficulty_performance}
\setlength{\tabcolsep}{1.25pt}
\renewcommand{\arraystretch}{1.1}
\resizebox{\textwidth}{!}{
\begin{tabular}{l|l|ccc|ccc|ccc|ccc|>{\columncolor{overallcolor}[\tabcolsep][\tabcolsep]}c>{\columncolor{overallcolor}[\tabcolsep][\tabcolsep]}c>{\columncolor{overallcolor}[\tabcolsep][\tabcolsep]}c}

\toprule
\multirow{2}{*}{Category} 
& \multirow{2}{*}{Method} 
& \multicolumn{3}{c|}{\begin{tabular}[c]{@{}c@{}}Domain Knowledge\\Reasoning\end{tabular}} 
& \multicolumn{3}{c|}{\begin{tabular}[c]{@{}c@{}}Empirical Discovery\\\& Simulation\end{tabular}} 
& \multicolumn{3}{c|}{\begin{tabular}[c]{@{}c@{}}Procedural Task\\Execution\end{tabular}} 
& \multicolumn{3}{c|}{\begin{tabular}[c]{@{}c@{}}Rule System\\Application\end{tabular}} 
& \multicolumn{3}{c}{\raisebox{0.55\baselineskip}{Overall}} \\
\cmidrule(lr){3-5}
\cmidrule(lr){6-8}
\cmidrule(lr){9-11}
\cmidrule(lr){12-14}
\cmidrule(lr){15-17}
& & Easy & Medium & Hard
& Easy & Medium & Hard
& Easy & Medium & Hard
& Easy & Medium & Hard
& \multicolumn{1}{c}{Easy}
& \multicolumn{1}{c}{Medium}
& \multicolumn{1}{c}{Hard} \\
\midrule

\multirow{3}{*}{Long-Context LLM}
& Gemini-3-Flash 
& 44.4 & 17.9 & 9.1
& 0.0 & 0.0 & 11.1
& 42.9 & 12.6 & 5.0
& 61.8 & 21.2 & 6.0
& 37.3 & 13.0 & 7.8 \\
& GPT-5-mini 
& 61.7 & \textbf{37.6} & \textbf{16.5}
& \textbf{50.0} & 0.0 & \textbf{22.2}
& 54.7 & \textbf{22.3} & \textbf{8.8}
& \textbf{76.7} & \textbf{26.6} & \textbf{9.9}
& \textbf{60.8} & \textbf{21.6} & \textbf{14.4} \\
\rowcolor{headergray}
\multicolumn{1}{l|}{\cellcolor{white}} & DeepSeek-V3.2 
& \textbf{64.6} & 21.4 & 0.0
& 10.0 & 0.0 & 0.0
& \textbf{57.9} & 10.0 & 0.0
& 75.7 & 18.8 & 0.0
& \cellcolor{overallcolor}52.1 
& \cellcolor{overallcolor}12.6 
& \cellcolor{overallcolor}0.0 \\

\specialrule{0.4pt}{1pt}{2pt}
\specialrule{0.4pt}{0pt}{3pt}

\multirow{3}{*}{\begin{tabular}[c]{@{}l@{}}Retrieval-Augmented\\Memory\end{tabular}}
& BM25 
& 60.9 & 23.5 & 4.8
& 0.0 & 0.0 & \underline{22.2}
& 54.5 & 13.2 & \textbf{12.1}
& \underline{73.5} & 18.4 & 1.4
& 47.2 & 13.8 & \underline{10.1} \\
& Qwen3-Emb-4B 
& \underline{63.5} & 23.7 & 6.7
& \textbf{10.0} & 0.0 & 11.1
& 56.3 & 13.0 & 8.1
& 70.3 & 18.3 & 5.9
& \textbf{50.0} & 13.7 & 7.9 \\
& GraphRAG 
& 60.4 & \underline{25.4} & \underline{7.6}
& 0.0 & 0.0 & 11.1
& \underline{59.2} & 13.4 & 9.2
& 72.4 & 15.9 & 2.8
& 48.0 & 13.7 & 7.7 \\

\midrule
\multirow{4}{*}{\begin{tabular}[c]{@{}l@{}}General Long-Term\\Memory\end{tabular}}
& Mem0 
& 55.3 & 15.9 & 5.1
& 0.0 & 0.0 & 0.0
& 50.5 & 12.5 & 4.4
& 72.6 & 18.4 & 5.8
& 44.6 & 11.7 & 3.8 \\
& A-MEM 
& 49.4 & 13.8 & 2.9
& \textbf{10.0} & 0.0 & 11.1
& 50.4 & 16.1 & 7.4
& 65.3 & 10.1 & 5.5
& 43.8 & 10.0 & 6.7 \\
& MemOS 
& 55.7 & \underline{25.4} & 4.8
& \textbf{10.0} & 0.0 & 11.1
& 57.5 & 16.3 & 7.4
& 63.9 & 17.3 & 4.0
& 46.8 & 14.8 & 6.8 \\
& MemoryOS 
& 51.5 & 19.0 & 1.9
& \textbf{10.0} & 0.0 & \underline{22.2}
& 52.7 & 12.4 & 6.6
& 62.7 & 12.7 & 5.4
& 44.2 & 11.0 & 9.0 \\

\midrule
\multirow{5}{*}{\begin{tabular}[c]{@{}l@{}}Procedural Long-Term\\Memory\end{tabular}}
& AWM 
& 54.2 & 22.7 & \textbf{9.0}
& 0.0 & 0.0 & 0.0
& \textbf{60.9} & 12.8 & 6.9
& 68.8 & 14.4 & 3.3
& 46.0 & 12.5 & 4.8 \\
& SkillWeaver 
& \textbf{64.4} & 23.5 & 4.8
& \textbf{10.0} & 0.0 & 11.1
& 58.8 & 7.1 & 2.2
& 60.5 & 14.8 & 4.5
& \underline{48.4} & 11.3 & 5.6 \\
& AgentKB 
& 56.9 & 22.1 & 5.1
& \textbf{10.0} & 0.0 & 0.0
& 51.2 & 10.3 & 5.5
& 64.5 & 12.7 & 3.7
& 45.6 & 11.3 & 3.6 \\
& ACE 
& 58.0 & 24.6 & 2.9
& 0.0 & 0.0 & \textbf{33.3}
& 46.2 & \underline{17.2} & 8.1
& \textbf{74.9} & \textbf{24.0} & \textbf{7.8}
& 44.8 & \textbf{16.5} & \textbf{13.0} \\
& ReasoningBank 
& 60.9 & \textbf{27.5} & 6.1
& 0.0 & 0.0 & 0.0
& 48.0 & \textbf{18.9} & 11.0
& 62.1 & 16.5 & \underline{7.4}
& 42.7 & \underline{15.7} & 6.1 \\

\midrule
\begin{tabular}[c]{@{}l@{}}Meta-Evolution\\Memory\end{tabular}
& MemEvolve 
& 54.3 & 16.7 & 3.8
& \textbf{10.0} & 0.0 & 11.1
& 47.3 & 12.4 & \underline{11.4}
& 72.5 & \underline{18.9} & 3.1
& 46.0 & 12.0 & 7.4 \\

\bottomrule
\end{tabular}
}
\end{table*}

\paragraph{Finding 5: Memory helps more on difficult tasks, but can hurt easy tasks.} Cross-episode knowledge evolution does not show uniform gains from memory.
As reported in Table~\ref{tab:task_type_difficulty_performance}, on the overall easy split, the memory-free DeepSeek-V3.2 achieves $52.1$, while the best memory method, Qwen3-Emb-4B, reaches only $50.0$.
General long-term memory methods perform even lower, ranging from $43.8$ to $46.8$.
In contrast, memory becomes more useful on harder tasks.
On the overall hard split, DeepSeek-V3.2 obtains $0.0$, while ACE improves the score to $13.0$, followed by BM25 with $10.1$ and MemoryOS with $9.0$.
This suggests that when the current context is already sufficient for the backbone model, additional memory may introduce irrelevant evidence, mismatched past information, or unnecessary prompt bias.

\paragraph{Finding 6: The bottleneck is forming reusable knowledge, not only storing more information.}  Table~\ref{tab:task_type_difficulty_performance} shows that
ACE achieves the best overall performance among memory methods on both medium and hard tasks, with $16.5$ and $13.0$, respectively.
Its advantage is especially clear on rule system application, where it obtains $74.9$, $24.0$, and $7.8$ on the easy, medium, and hard splits, outperforming all other memory methods at each difficulty level.
This indicates that effective memory should support how knowledge is applied, rather than only what knowledge is stored.

\subsubsection{Cross-Episode Execution Evolution}

\newcommand{\numfont}{\fontsize{12pt}{15pt}\selectfont}
\newcommand{\groupfont}{\fontsize{12pt}{14pt}\selectfont}
\newcommand{\headfont}{\normalsize}
\newcolumntype{N}{>{\numfont}c}
\newcolumntype{A}{>{\columncolor{avgcolor}[\tabcolsep][\tabcolsep]\numfont}c}
\newcolumntype{O}{>{\columncolor{overallcolor}[\tabcolsep][\tabcolsep]\numfont}c}

\begin{table*}[htbp]
\centering
\caption{Success rate (\%) comparison across \textsc{CrossEp-Tool}, \textsc{CrossEp-Web} and \textsc{CrossEp-Emb}. Best results among memory methods are bolded, and runner-up results are underlined. All memory-augmented methods use DeepSeek-V3.2 as the backbone.}
\label{tab:overall_rank_benchmark}
\setlength{\tabcolsep}{2pt}
\renewcommand{\arraystretch}{1.15}
\resizebox{\textwidth}{!}{
\begin{threeparttable}
\begin{tabular}{l|NNNNA|NNA|NNNNNNA|O}

\toprule
\multirow{2}{*}[-1.5ex]{Method}
& \multicolumn{5}{c|}{Tool Using}
& \multicolumn{3}{c|}{Web Search}
& \multicolumn{7}{c|}{Embodied AI}
& \multicolumn{1}{c}{\multirow{2}{*}[-1.5ex]{\headfont{\makecell[c]{Overall\\Rank}}}} \\
\cmidrule(lr){2-6}
\cmidrule(lr){7-9}
\cmidrule(lr){10-16}
& \headfont{\makecell[c]{Gorilla File\\System}}
& \headfont{\makecell[c]{Vehicle\\Control}}
& \headfont{\makecell[c]{Trading\\Bots}}
& \headfont{\makecell[c]{Travel\\Booking}}
& \multicolumn{1}{c|}{\headfont{\makecell[c]{Avg.\\Rank}}}
& \headfont{\makecell[c]{xbench\\-DS}}
& \headfont{\makecell[c]{WebWalker\\QA}}
& \multicolumn{1}{c|}{\headfont{\makecell[c]{Avg.\\Rank}}}
& \headfont{\makecell[c]{Examine\\in Light}}
& \headfont{\makecell[c]{Pick\\\&Place}}
& \headfont{\makecell[c]{Clean\\\&Place}}
& \headfont{\makecell[c]{Cool\\\&Place}}
& \headfont{\makecell[c]{Heat\\\&Place}}
& \headfont{\makecell[c]{Pick Two\\\&Place}}
& \multicolumn{1}{c|}{\headfont{\makecell[c]{Avg.\\Rank}}}
& \multicolumn{1}{c}{} \\
\midrule

\multicolumn{17}{c}{\groupfont\emph{Long-Context LLM}} \\[2pt]
Gemini-3-Flash
& \textbf{74.0} & \textbf{62.0} & \textbf{88.0} & \textbf{54.0} & \textbf{3.0}
& \textbf{76.0} & \textbf{72.4} & \textbf{5.0}
& 68.4 & 76.1 & 43.2 & \textbf{64.3} & \textbf{60.0} & 26.7 & 12.0
& \textbf{7.8} \\
GPT-5-mini
& 50.0 & 50.0 & 72.0 & 44.0 & 9.5
& 72.0 & 71.8 & 10.0
& 36.8 & 69.6 & 21.6 & 25.0 & 40.0 & 40.0 & 15.5
& 12.7 \\
\rowcolor{headergray}
DeepSeek-V3.2
& 64.0 & 36.0 & 54.0 & 26.0 & \cellcolor{avgcolor}13.5
& 74.0 & 61.2 & \cellcolor{avgcolor}14.0
& \textbf{73.7} & \textbf{93.5} & \textbf{56.8} & 60.7 & 56.0 & \textbf{57.8} & \cellcolor{avgcolor}\textbf{10.0}
& \cellcolor{overallcolor}11.8 \\

\specialrule{0.4pt}{1pt}{2pt}
\specialrule{0.4pt}{0pt}{3pt}

\multicolumn{17}{c}{\groupfont\emph{Retrieval-Augmented Memory}} \\[3pt]
BM25
& 68.0 & \textbf{88.0} & 66.0 & 32.0 & 5.8
& \underline{79.0} & 65.9 & 6.5
& 68.4 & \textbf{97.8} & 62.2 & 60.7 & 64.0 & 71.1 & 7.8
& 6.9 \\
Qwen3-Emb-4B
& 68.0 & \underline{82.0} & \underline{78.0} & 34.0 & \textbf{4.0}
& 78.0 & 65.3 & 8.0
& \underline{79.0} & \textbf{97.8} & 70.3 & 60.7 & 64.0 & 71.1 & 6.0
& 5.7 \\
GraphRAG
& 66.0 & 68.0 & 72.0 & \underline{40.0} & 6.5
& 73.0 & 70.0 & 11.0
& 63.2 & \textbf{97.8} & 75.7 & 60.7 & 72.0 & 77.8 & 5.8
& 6.9 \\

\addlinespace[3pt]
\multicolumn{17}{c}{\groupfont\emph{General Long-Term Memory}} \\[3pt]
Mem0
& \textbf{76.0} & 58.0 & 48.0 & 30.0 & 8.5
& \textbf{82.0} & 70.6 & \textbf{4.0}
& \underline{79.0} & 89.1 & 67.6 & 57.1 & 68.0 & 53.3 & 8.2
& 7.9 \\
A-MEM
& 68.0 & 62.0 & 70.0 & 26.0 & 8.8
& 76.0 & 71.8 & 6.0
& \textbf{84.2} & 91.9 & \textbf{81.1} & 64.3 & \textbf{80.0} & \textbf{84.4} & \textbf{1.8}
& \textbf{5.2} \\
MemOS
& \underline{72.0} & 60.0 & 36.0 & 28.0 & 9.8
& 78.0 & 71.2 & 5.0
& 63.2 & 91.3 & 67.6 & 64.3 & 68.0 & 53.3 & 7.8
& 8.4 \\
MemoryOS
& \underline{72.0} & 72.0 & 56.0 & 36.0 & 5.8
& 78.0 & 69.4 & 6.5
& 73.7 & \underline{95.7} & 75.7 & 64.3 & 60.0 & 57.8 & 6.5
& 6.5 \\

\addlinespace[3pt]
\multicolumn{17}{c}{\groupfont\emph{Procedural Long-Term Memory}} \\[3pt]
AWM
& 64.0 & 56.0 & 74.0 & 24.0 & 11.0
& 76.0 & \underline{72.4} & 5.0
& 73.7 & 91.3 & \underline{78.4} & 71.4 & \underline{76.0} & 64.4 & 4.3
& 7.2 \\
SkillWeaver
& 66.0 & 48.0 & 44.0 & 28.0 & 12.3
& 74.0 & 64.7 & 13.5
& 63.2 & 84.8 & 59.5 & 53.6 & 64.0 & 66.7 & 10.2
& 12.2 \\
AgentKB
& 62.0 & 71.4 & 74.0 & 30.0 & 8.3
& \underline{79.0} & 68.8 & 6.0
& 68.4 & 89.1 & 67.6 & \textbf{85.7} & \textbf{80.0} & 73.3 & \underline{4.2}
& 6.5 \\
ACE
& 68.0 & 56.0 & 14.0 & 34.0 & 9.8
& 77.0 & 65.3 & 9.5
& \textbf{84.2} & 78.3 & 75.7 & 21.4 & \underline{76.0} & \underline{82.2} & 4.8
& 8.2 \\
ReasoningBank
& 64.0 & \underline{82.0} & \textbf{86.0} & \textbf{42.0} & \underline{4.8}
& 76.0 & \textbf{75.3} & \underline{4.5}
& \underline{79.0} & 87.0 & \textbf{81.1} & 67.9 & 60.0 & 73.3 & 4.5
& \underline{5.4} \\

\addlinespace[3pt]
\multicolumn{17}{c}{\groupfont\emph{Meta-Evolution Memory}} \\[3pt]
MemEvolve
& 66.0 & 70.0 & 66.0 & 28.0 & 8.8
& 73.0 & 65.3 & 13.0
& 57.9 & 89.1 & 64.9 & \underline{78.6} & \underline{76.0} & 75.6 & 6.0
& 8.8 \\

\bottomrule
\end{tabular}
\end{threeparttable}
}
\end{table*}

\paragraph{Finding 7: Different execution domains favor different memory forms.}
Table~\ref{tab:overall_rank_benchmark} shows that the best memory family changes across execution domains.
In Tool Using, retrieval-augmented memory achieves the best family-level average rank among memory methods, with an average rank of $5.43$.
In Web Search, general long-term memory performs best, with an average rank of $5.38$.
In Embodied AI, procedural long-term memory performs best, with an average rank of $5.60$.
This result shows that cross-episode execution evolution cannot be addressed by one fixed memory form.
Retrieval-augmented memory reuses similar past cases, which fits tool-use tasks where similar API calls and parameter patterns recur.
General long-term memory maintains persistent information, which is useful in web search where evidence and verification habits must be accumulated.
Procedural long-term memory stores strategies, workflows, or skills, which fits embodied tasks where action routines and affordances are important.

\begin{figure}[htbp]
    \centering
    \includegraphics[width=1\linewidth]{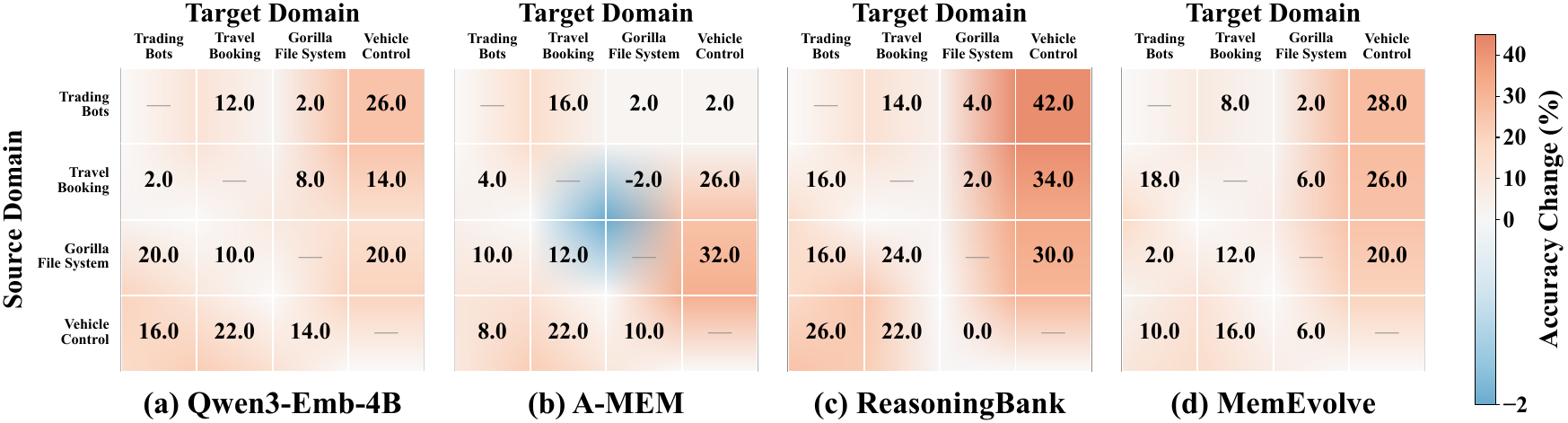}
    \caption{Cross-environment transfer results in \textsc{CrossEp-Tool}.}
    \label{fig:tool_using_heatmap}
\end{figure}

\begin{figure}[htbp]
    \centering
    \includegraphics[width=1\linewidth]{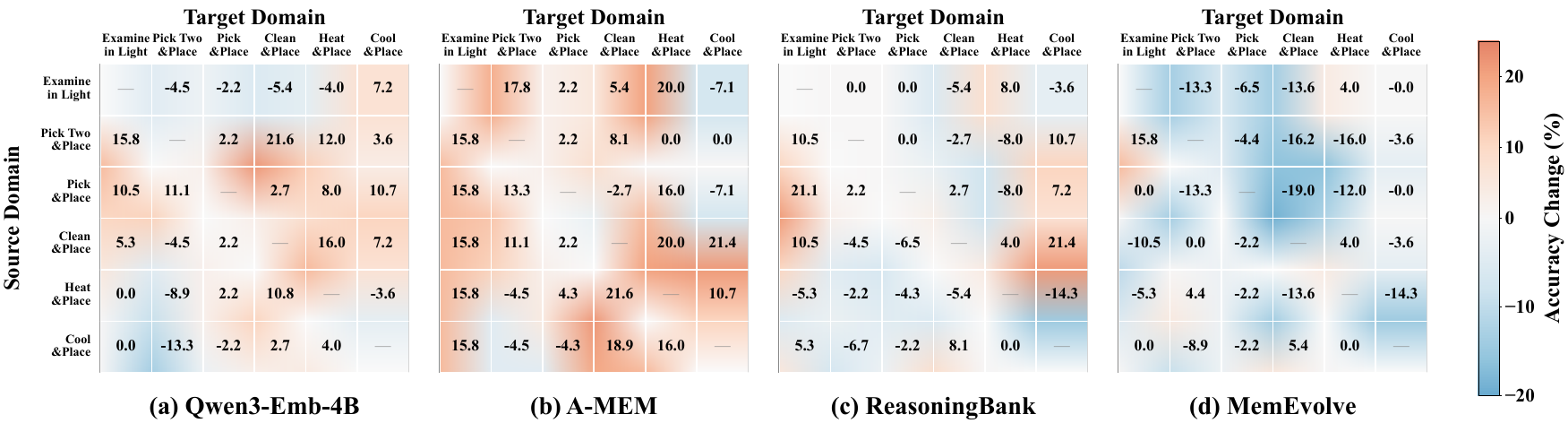}
    \caption{Cross-environment transfer results in \textsc{CrossEp-Emb.}}
    \label{fig:embody_ai_heatmap}
\end{figure}

\paragraph{Finding 8: Transfer depends on whether memory matches the target decision process.}
Figure~\ref{fig:tool_using_heatmap} indicates that tool-use tasks exhibit more stable positive transfer. 
This pattern may be related to the fact that different tool-use environments often involve similar decision steps, such as checking intermediate states, using tool outputs to choose the next call, and verifying the final state. 
Memory that preserves such decision procedures may therefore be more reusable across tool-use domains.
In contrast, Figure~\ref{fig:embody_ai_heatmap} shows that transfer in ALFWorld is less stable and more likely to become neutral or negative. 
Compared with tool-use domains, ALFWorld sub-environments differ more substantially in their task requirements and execution conditions.
As a result, experience accumulated in one embodied setting may not apply as consistently to another. 
Overall, these results suggest that cross-environment transfer depends not only on whether memory stores past execution experience, but also on whether that experience remains aligned with the target decision process.

\subsubsection{Summary}
These findings reveal a consistent pattern. (1) Current memory methods can improve agent performance, but not uniformly: strong memory-free long-context baselines remain highly competitive, and explicit memory provides the clearest gains only in some settings. (2) Memory helps most when the current context is insufficient, the task is more difficult, or the stored experience matches the target decision process; it can hurt when memory introduces irrelevant evidence, removes fine-grained execution details through compression, or transfers mismatched procedures across environments. (3) No single memory form is sufficient across all settings: retrieval remains a strong choice for knowledge-oriented demands, while procedural and long-term memory are more effective for execution-oriented problems when their stored form aligns with the reusable structure of the task. Overall, current methods still do not provide a general solution for self-evolving agent memory.

\section{Conclusion and Future Works}
In this work, we introduce \EMB, a benchmark for evaluating how agent memory evolves within and across episodes. \EMB~ covers four settings by combining two axes: memory scope (in-episode vs.~cross-episode) and memory content (knowledge-oriented vs.~execution-oriented), providing a unified testbed for adaptive agent memory. Experiments show that current memory methods are still far from a general solution. We hope \EMB~ will support future research on more effective memory systems for LLM-based agents.

\paragraph{Limitations}
EvoMemBench is designed as an evaluation benchmark for existing agent-memory methods rather than a new memory architecture, and therefore does not directly propose solutions to the weaknesses identified through its evaluation. Specifically, current methods remain far from a general solution: strong long-context baselines are still highly competitive, and no single memory form works consistently across all settings. These findings motivate future work on selectively using memory when beneficial, and adapting memory forms to the reusable structure of each task.

\section*{Acknowledgement}
This research was supported by NSFC 62572418, the Guangdong
Provincial Talent Program under Grant (No.2024TQ08X366), and HKUST-GZ Create Link Joint Lab. This work was partially supported by the Research Travel Grant of the Base of Red Bird MPhil (RBM) at The Hong Kong University of Science and Technology (Guangzhou). We would like to thank our RBM Project Supervisor Zheng Luo for the academic support.
\bibliographystyle{plainnat}
\bibliography{reference}

\clearpage
\appendix
\section{Additional Details on \EMB}
\subsection{Details of Datasets}
\label{subapp:datasets}

All datasets used in our study are previously published benchmarks covering diverse domains including context learning, tool-use evaluation, memory-intensive agent interaction, profession-aligned real-world tasks, web navigation, and embodied decision-making environments. The detailed descriptions of these datasets and their associated evaluation metrics are as follows:

\begin{itemize}[leftmargin=*, itemsep=2pt, topsep=2pt]

\item \textbf{CL-Bench.} The CL-Bench \cite{dou2026cl} dataset is a comprehensive benchmark designed to evaluate language models’ context learning capability, i.e., the ability to acquire and apply new knowledge provided within the context rather than relying solely on pre-trained knowledge. It consists of 500 complex contexts, 1,899 tasks, and more than 31K verification rubrics spanning 4 major categories and 18 subcategories, including domain knowledge reasoning, procedural task execution, and empirical discovery. Unlike traditional long-context benchmarks that primarily focus on retrieval or reading comprehension, CL-Bench requires models to genuinely learn novel knowledge and rules from context. Model performance is evaluated using task-solving accuracy. In our experiments, CL-Bench is used to evaluate long-context memory retention and context-dependent reasoning ability.

\item \textbf{BFCL-V3-MultiTurn.} The BFCL benchmark \cite{patil2025bfcl} is a large-scale evaluation suite for assessing tool-use and function-calling capabilities of large language models. The BFCL-V3-MultiTurn setting extends earlier versions by introducing multi-turn interactions, multi-step tool invocation, and parameter completion scenarios, enabling more realistic evaluation of agentic behavior in complex environments. The benchmark evaluates whether models can correctly select APIs, generate executable arguments, and maintain interaction consistency across multiple turns. Performance is primarily measured using function-calling accuracy and execution success rate. 

\item \textbf{MemoryAgentBench.} The MemoryAgentBench \cite{hu2025evaluating} dataset is a benchmark specifically designed for evaluating memory capabilities in LLM-based agents through incremental multi-turn interactions. The benchmark identifies four core competencies essential for memory agents, including accurate retrieval, test-time learning, long-range understanding, and selective forgetting. To simulate realistic conversational settings, the benchmark reformulates existing long-context datasets into incremental multi-turn interactions and introduces newly constructed tasks such as EventQA and FactConsolidation. Model performance is evaluated using memory retrieval accuracy and downstream task completion accuracy under progressively increasing interaction lengths. 

\item \textbf{xBench.} The xBench \cite{chen2025xbench} benchmark is a profession-aligned real-world evaluation suite designed to measure productivity scaling behavior in autonomous agents. It contains realistic long-horizon tasks derived from professional workflows, including planning, coding, document processing, information analysis, and web interaction. Compared with conventional synthetic benchmarks, xBench emphasizes realistic task execution and environment interaction under practical constraints. Model performance is evaluated based on task success rate and execution quality across diverse professional domains. 

\item \textbf{WebWalkerQA.} The WebWalkerQA benchmark \cite{wu2025webwalker} is a web traversal and multi-hop web reasoning benchmark designed to evaluate LLM agents’ capability in navigating real-world websites and extracting information across multiple webpages. The associated WebWalkerQA task requires agents to sequentially browse webpages, collect evidence from different sources, and answer complex multi-hop questions. The benchmark evaluates both navigation success and final question-answering accuracy, making it suitable for assessing long-horizon planning, memory retention, and web interaction capabilities.

\item \textbf{ALFWorld.} The ALFWorld \cite{shridhar2021alfworld} benchmark is a text-based embodied environment that aligns textual interaction with embodied household environments. Built upon the ALFRED embodied instruction-following benchmark, ALFWorld converts embodied tasks into text-based interactive environments involving navigation, object manipulation, and sequential decision-making in partially observable settings. Tasks typically require long-horizon planning and persistent memory of previously observed states. Model performance is evaluated using task success rate and goal completion accuracy. 

\end{itemize}

\subsection{Details of Metrics}
\label{app:metric_details}

We evaluate memory methods from two complementary perspectives: task performance and efficiency.
Different metrics are used depending on the type of evaluation setting.

\paragraph{Answer Accuracy.}
For knowledge-oriented settings, including \textsc{InEp-Know} and \textsc{CrossEp-Know}, we report \textit{answer accuracy}, defined as the proportion of evaluation instances answered correctly:
\[
\mathrm{Acc}
=
\frac{1}{N}
\sum_{i=1}^{N}
\mathrm{Correct}(i),
\]
where $\mathrm{Correct}(i)\in\{0,1\}$ indicates whether the prediction for instance $i$ is correct.

\paragraph{Success Rate.}
For execution-oriented settings, including \textsc{InEp-Exec}, \textsc{CrossEp-Tool}, \textsc{CrossEp-Web}, and \textsc{CrossEp-Emb}, we report \textit{success rate}.
Let $\mathrm{Success}(i)\in\{0,1\}$ indicate whether the agent successfully completes the final goal of task $i$ according to the task-specific evaluator.
The success rate is defined as
\[
\mathrm{SR}
=
\frac{1}{N}
\sum_{i=1}^{N}
\mathrm{Success}(i).
\]
This metric captures complete task accomplishment and is therefore a strict measure of end-to-end execution quality.

\paragraph{Progress Score.}
In \textsc{InEp-Exec} and \textsc{CrossEp-Tool}, we additionally report \textit{progress score} to characterize partial task completion.
These tasks involve multi-turn tool-use trajectories in which an agent may correctly complete earlier turns or intermediate requirements even if it eventually fails to solve the full task.
For task $i$, let $T_i$ denote the number of evaluated turns or execution checkpoints, and let $\mathrm{Pass}(i,t)\in\{0,1\}$ indicate whether the agent satisfies the evaluator at checkpoint $t$.
The task-level progress score is
\[
\mathrm{PS}_i
=
\frac{1}{T_i}
\sum_{t=1}^{T_i}
\mathrm{Pass}(i,t),
\]
and the overall progress score is
\[
\mathrm{PS}
=
\frac{1}{N}
\sum_{i=1}^{N}
\mathrm{PS}_i.
\]
Unlike success rate, which only rewards fully completed tasks, progress score provides a finer-grained view of how far an agent proceeds during execution.

\paragraph{Average Steps.}
We report \textit{average steps} for \textsc{InEp-Exec}, \textsc{CrossEp-Tool}, \textsc{CrossEp-Web}, and \textsc{CrossEp-Emb}.
For each task $i$, let $\mathrm{Steps}(i)$ denote the number of agent execution steps taken during the trajectory, where a step follows the native interaction unit of the corresponding environment or agent framework.
Average steps is computed as
\[
\mathrm{AS}
=
\frac{1}{N}
\sum_{i=1}^{N}
\mathrm{Steps}(i).
\]
This metric reflects interaction efficiency: lower values indicate that the agent reaches termination with fewer execution steps.
We report it together with success rate to distinguish methods that are merely more successful from those that are both successful and efficient.

\paragraph{Token Usage.}
We use \textit{token usage} as a general efficiency metric.
For each evaluation instance, token usage includes all LLM inference tokens consumed by both the task agent and the memory module, covering input and output tokens across all model calls.
Let $\mathcal{C}_i$ denote the set of LLM calls made for instance $i$, and let $\mathrm{Tok}^{\mathrm{in}}(c)$ and $\mathrm{Tok}^{\mathrm{out}}(c)$ denote the input and output tokens of call $c$.
The token usage for instance $i$ is
\[
\mathrm{TU}_i
=
\sum_{c\in\mathcal{C}_i}
\left(
\mathrm{Tok}^{\mathrm{in}}(c)
+
\mathrm{Tok}^{\mathrm{out}}(c)
\right).
\]
We report token usage to quantify the computational overhead introduced by different memory mechanisms.

\subsection{Details of Baselines}
\label{subapp:baselines}

We compare our framework against representative baselines spanning six categories, including long-context LLMs, retrieval-augmented memory systems, short-term memory architectures, general long-term memory systems, procedural long-term memory methods, and meta-evolution memory frameworks. Below we provide detailed descriptions of these baselines.

\begin{itemize}[leftmargin=*, itemsep=2pt, topsep=2pt]

\item \textbf{Long-Context LLMs.}
\begin{itemize}

\item \textbf{Gemini-3-Flash} \cite{gemini3flash} is a proprietary long-context language model developed by Google. It supports efficient processing of extremely long contexts and demonstrates strong performance on retrieval, reasoning, and multimodal understanding tasks. Its optimized sparse attention and long-context inference mechanisms make it a strong baseline for large-scale memory-intensive applications. Its efficient long-context decoding strategy also enables stable performance under extremely large context windows with relatively low inference latency.

\item \textbf{GPT-5-mini} \cite{openai2025gpt5} is a lightweight variant of OpenAI’s GPT-5 family optimized for efficient inference and long-context interaction. Despite its compact size, it maintains strong instruction-following and reasoning abilities, serving as a representative efficient long-context proprietary model. The model is particularly suitable for interactive agent settings requiring fast response generation under constrained computational budgets.

\item \textbf{DeepSeek-V3.2} \cite{liu2025deepseek} is an advanced mixture-of-experts (MoE) language model developed by DeepSeek AI. It incorporates efficient routing mechanisms and large-scale continual pretraining to support long-context reasoning, coding, and retrieval-intensive tasks with strong scalability. Its MoE architecture allows scalable activation of specialized experts, improving efficiency for heterogeneous reasoning and retrieval workloads.

\end{itemize}

\item \textbf{Retrieval-Augmented Memory.}
\begin{itemize}

\item \textbf{BM25} \cite{robertson2009probabilistic} is a classical sparse retrieval algorithm based on term-frequency and inverse document frequency statistics. It remains a strong lexical retrieval baseline for long-context question answering and memory retrieval tasks. Due to its exact lexical matching behavior, BM25 is particularly robust for retrieving sparse factual memories and keyword-centric information.

\item \textbf{Qwen3-Emb-4B} \cite{zhang2025qwen3} is a 4B parameter embedding model built upon the Qwen3 foundation model. It adopts instruction-aware embedding training and supports multilingual semantic retrieval, providing strong dense retrieval capability for memory-intensive agent systems. Its strong semantic embedding capability enables effective retrieval even when queries and memories exhibit substantial lexical variation.

\item \textbf{GraphRAG} \cite{edge2024local} is a graph-based retrieval-augmented generation framework that first constructs an entity-centric knowledge graph from source documents, then performs hierarchical community summarization for query answering. Compared with conventional vector RAG systems, GraphRAG is particularly effective for corpus-level reasoning and global summarization tasks. The graph-based organization additionally improves multi-hop retrieval and relation-aware reasoning across large document collections.

\end{itemize}

\item \textbf{Short-Term Memory.}
\begin{itemize}

\item \textbf{MemAgent} \cite{yu2025memagent} reformulates long-context modeling as a reinforcement learning problem with a fixed-size memory buffer. The model iteratively reads document chunks and updates compressed memory representations through overwrite operations. It employs Multi-Conversation DAPO training and achieves strong extrapolation ability under linear computational complexity. The overwrite-based memory update mechanism enables bounded memory usage while maintaining long-context reasoning capability.

\item \textbf{MemoBrain} \cite{qian2026memobrain} is a short-term memory framework designed for dynamic conversational reasoning. It continuously compresses recent interaction histories into compact memory states, enabling efficient context maintenance and rapid memory updates during agent interaction. Its executive-style memory abstraction mechanism is particularly designed to support dynamic reasoning over evolving conversational contexts.

\end{itemize}

\item \textbf{General Long-Term Memory.}
\begin{itemize}

\item \textbf{Mem0} \cite{chhikara2025mem0} is a scalable long-term memory architecture for conversational agents. It incrementally extracts salient information from interactions and manages memory using four operations: \texttt{ADD}, \texttt{UPDATE}, \texttt{DELETE}, and \texttt{NOOP}. The framework also introduces graph-based memory representations for enhanced relational reasoning. Its fine-grained memory operation design allows flexible incremental memory maintenance across long-term interactions.

\item \textbf{A-MEM} \cite{xu2025mem} is a structured memory framework that organizes historical interactions into interconnected semantic notes. It emphasizes adaptive memory organization and long-term retrieval for persistent agent reasoning across sessions. The framework further supports memory association and semantic linking across temporally distant interactions.

\item \textbf{MemOS} \cite{li2025memos} is an operating-system-inspired memory management framework for AI agents. It adopts hierarchical storage and segmented memory management strategies to coordinate memory allocation, retrieval, and long-term persistence across complex interactions. Its operating-system-inspired abstraction enables modular coordination between memory scheduling, retrieval, and persistence modules.

\item \textbf{MemoryOS} \cite{kang2025memory} introduces a three-tier memory hierarchy consisting of short-term memory (STM), mid-term memory (MTM), and long-term personal memory (LPM). The framework integrates memory storage, updating, retrieval, and response generation modules, enabling dynamic conversational memory evolution and long-term personalization. The hierarchical memory design enables adaptive retention of both recent interaction context and long-term personalized knowledge.

\end{itemize}

\item \textbf{Procedural Long-Term Memory.}
\begin{itemize}

\item \textbf{AWM (Agent Workflow Memory)} \cite{wang2025awm} enables agents to induce reusable workflows from historical trajectories and store them as structured memory units. During inference, the agent retrieves relevant workflows to guide future planning and execution. AWM supports both offline workflow induction from demonstrations and online induction from self-generated successful trajectories. By reusing previously induced workflows, AWM improves planning efficiency and reduces redundant exploration during agent execution.

\item \textbf{SkillWeaver} \cite{zheng2025skillweaver} is a self-improving web-agent framework that autonomously discovers, synthesizes, and refines reusable skills represented as executable APIs. It consists of three stages: skill proposal, skill synthesis, and skill honing, enabling continuous procedural skill accumulation through environment interaction. The framework enables continual procedural self-improvement through autonomous skill refinement across repeated web interactions.

\item \textbf{AgentKB} \cite{tang2025agent} is a cross-framework knowledge-sharing infrastructure for heterogeneous agent systems. It abstracts execution trajectories into reusable experience units and employs hybrid retrieval together with disagreement-gated integration to enable stable cross-agent knowledge transfer.

\item \textbf{ACE} \cite{zhang2026agentic} is a procedural memory framework that accumulates executable agent experiences as reusable action templates. It focuses on experience abstraction and adaptive execution reuse for complex sequential decision-making tasks. The framework is particularly effective for repeated sequential tasks that benefit from reusable execution patterns.

\item \textbf{ReasoningBank} \cite{ouyang2026reasoningbank} is a reasoning-oriented memory repository that stores structured reasoning traces and intermediate decision trajectories. It enhances long-horizon planning by retrieving relevant reasoning patterns from historical problem-solving experiences. The stored reasoning trajectories additionally support retrieval-based reasoning generalization across related problem-solving tasks.

\end{itemize}

\item \textbf{Meta-Evolution Memory.}
\begin{itemize}

\item \textbf{MemEvolve} \cite{zhang2025memevolve} is a meta-evolutionary memory framework that continuously refines and evolves stored memories through iterative self-optimization. The framework dynamically updates memory structures based on long-term task feedback, enabling adaptive memory evolution and continual learning. Its self-evolution mechanism enables continuous adaptation of memory structures under changing task distributions and environments.

\end{itemize}

\end{itemize}

Notably, since different settings expose different memory interfaces, we evaluate only applicable methods in each setting: procedural long-term memory is excluded from in-episode knowledge evolution, and short-term memory is excluded from cross-episode settings. 

\section{Additional Details on Implementations}
\label{app:implementation}
\subsection{Details of \textsc{INEP-EXEC} and \textsc{Cross-Tool} Construction}
\subsubsection{Prompt}
\begin{promptboxapp}
You are a dataset constructor for a multi-session agent-memory benchmark.

Your task is to jointly rewrite a multi-turn tool-use example by splitting any turn whose ground-truth contains multiple consecutive actions into multiple turns.

Goal:
Transform the original example into a more explicit multi-session sequence of dependent subtasks, while preserving the original semantics and increasing the need for cross-turn memory.

Core principle:
After splitting, if a later rewritten turn depends on an object, result, or state established in an earlier rewritten turn, that dependency should be expressed implicitly whenever possible, rather than by explicitly restating the exact entity name, ID, parameter value, or resolved result.

Rules:
- Only split a turn if its ground-truth contains multiple actions that form a meaningful sequential dependency.
- Split queries and split ground-truth must remain semantically aligned.
- The new ground-truth must be derivable from the old ground-truth by ordered slicing only.
- Do NOT invent new actions.
- Do NOT delete actions.
- Do NOT reorder actions.
- Do NOT add new constraints.
- Do NOT change action arguments unless the meaning stays exactly the same.

A turn is splittable only if:
1. Its ground-truth contains 2 or more actions; and
2. those actions form a natural user-level dependency, such as:
  - lookup -> use result
  - authenticate -> perform protected action
  - prepare state -> execute action
  - retrieve current info -> commit operation

Do NOT split:
- single-action turns
- turns whose actions are merely parallel
- turns whose internal steps are only low-level implementation details without meaningful subtask boundaries

Rewrite guidance:
- Each split query must sound natural as a user request.
- The first split query should request the first meaningful substep.
- Later split queries should naturally depend on earlier ones.
- Prefer implicit cross-turn references over explicit restatement.
- Preserve the original user intent and style.
- Keep the whole rewritten dialogue coherent.

Important implicit-reference rule:
When a later rewritten turn depends on something already established in an earlier rewritten turn, do NOT explicitly restate that earlier-established object if the dependency can be expressed naturally and unambiguously through an implicit reference.

Prefer references like:
- "the directory you created"
- "the one we just found"
- "that order"
- "those details"
- "the trip"
- "the message I sent"
- "the stock we just checked"
- "the reservation"
- "the result"

Avoid explicit restatement like:
- exact directory/file names already introduced in an earlier split turn
- exact IDs already resolved in an earlier split turn
- exact parameter values already determined in an earlier split turn
- exact entity names that should now be recoverable only from memory

However, do NOT make the query ambiguous or underspecified.
If an explicit mention is necessary to preserve clarity or avoid multiple valid interpretations, it may be kept.

What to keep explicit:
- Information newly introduced in the current rewritten turn
- Constraints that first appear in the current rewritten turn
- Information necessary to avoid ambiguity

What to make implicit when possible:
- Objects created earlier
- Results computed earlier
- Items selected earlier from a list
- State changes caused earlier
- Previously resolved destinations, bookings, tickets, orders, directories, messages, or chosen entities

Ground-truth handling:
- For a split turn, partition the original action list into consecutive slices.
- Each new query turn must align to exactly one consecutive slice.
- Concatenating all new slices for that original turn must exactly recover the original action list.

Hard constraints:
1. Never split a single-action turn.
2. Never merge two original turns.
3. Never split a turn into more parts than the number of actions in its original ground-truth.
4. Every split part must contain at least one action.
5. Later split queries may refer to earlier split queries, but must not depend on future turns.
6. Keep unsplit turns semantically unchanged except for light editing if needed for flow.
7. After splitting, do not expose cross-turn dependencies more explicitly than in the original task.
8. Prefer memory-dependent phrasing over answer-revealing phrasing.

Positive example:
Original turn:
Query: Move 'final_report.pdf' within document directory to 'temp' directory in document. Make sure to create the directory
Ground truth:
- cd(folder='document')
- mkdir(dir_name='temp')
- mv(source='final_report.pdf', destination='temp')

Good rewrite:
Turn 1
Query: Change to the document directory and create a 'temp' directory there.
Ground truth:
- cd(folder='document')
- mkdir(dir_name='temp')

Turn 2
Query: With that ready, move 'final_report.pdf' into the directory you created.
Ground truth:
- mv(source='final_report.pdf', destination='temp')

Negative rewrite:
Turn 1
Query: Change to the document directory and create a 'temp' directory there.
Ground truth:
- cd(folder='document')
- mkdir(dir_name='temp')

Turn 2
Query: With that ready, move 'final_report.pdf' into the 'temp' directory within document.
Ground truth:
- mv(source='final_report.pdf', destination='temp')

Why negative:
Because the second turn explicitly restates the key object ('temp' directory) that should instead be recoverable from memory.

Return JSON only in the format:
{
  "id": "<same id as input>",
  "rewritten_question": [
    [{"role": "user", "content": "..."}],
    ...
  ],
  "rewritten_ground_truth": [
    ["action1(...)"],
    ...
  ],
  "split_map": [
    {
      "original_turn": 1,
      "was_split": true,
      "new_turns": [1, 2]
    }
  ]
}

Before answering, internally verify:
1. Every original action appears exactly once in rewritten_ground_truth.
2. The global action order is preserved.
3. Each rewritten query matches its rewritten ground-truth.
4. No new actions were introduced.
5. No original actions were dropped.
6. The rewritten dialogue is coherent.
7. Later rewritten turns use implicit references wherever appropriate.
8. Cross-turn dependencies are not unnecessarily exposed by explicit restatement.

Now process the following input.

Query JSON:
<PASTE_QUERY_JSON_HERE>

Ground Truth JSON:
<PASTE_GROUND_TRUTH_JSON_HERE>
\end{promptboxapp}

\subsubsection{Human Validation Process}
\label{sec:human-validation}

The dataset constructed with the assistance of GPT-5-mini was validated by four annotators with experience in natural language processing and agent-memory tasks. Validation comprised two rounds of independent review. In the first round, each annotator reviewed 200 non-overlapping samples, covering all 800 samples. In the second round, every sample was reassigned to a different annotator, ensuring that each sample received two independent reviews. Each sample was evaluated against its original BFCL example according to the following criteria:
\begin{enumerate}
    \item \textbf{Content fidelity:} The rewritten dialogue preserves the original intent and constraints, without adding, omitting, modifying, duplicating, or reordering any action or argument.
    \item \textbf{Decomposition validity:} Decomposition is restricted to meaningfully dependent actions, and the resulting action slices reconstruct the original action sequence exactly.
    \item \textbf{Reference clarity:} Each implicit reference is unambiguous and can be resolved from the preceding dialogue turns.
\end{enumerate}

The two independent review outcomes were compared prior to adjudication. Any sample flagged by either reviewer was jointly re-examined, and samples requiring manual correction underwent an additional verification pass before acceptance. Overall, 780 samples (97.5\%) were accepted without revision, while 20 samples (2.5\%) were accepted after manual correction. Before adjudication, the two reviewers agreed on the overall validation decision for 775 of the 800 samples (96.9\%). Table~\ref{tab:human-validation} summarizes the validation statistics.

\begin{table}[htbp]
    \centering
    \caption{Human validation statistics for INEP-EXEC.}
    \label{tab:human-validation}
    \begin{tabularx}{\linewidth}{@{}Xr@{}}
        \toprule
        \textbf{Validation statistic} & \textbf{Value} \\
        \midrule
        Unique reconstructed samples & 800 \\
        Number of annotators & 4 \\
        First-round samples per annotator & 200 \\
        Independent reviews per sample & 2 \\
        Sample-level reviews across both rounds & 1{,}600 \\
        Accepted without revision & 780 (97.5\%) \\
        Accepted after manual correction & 20 (2.5\%) \\
        Pre-adjudication exact agreement on the overall decision
            & 775/800 (96.9\%) \\
        \bottomrule
    \end{tabularx}
\end{table}

\section{Additional Experimental Results}
\label{app:additional_results}

This section provides supplementary results for \textsc{InEp-Exec}, cross-environment transfer in cross-episode execution settings, and efficiency analysis in cross-episode settings.

\subsection{Additional Results for Cross-Environment Transfer in Cross-Episode Execution Settings}
\label{app:cross_environment_transfer}

We provide the full cross-environment transfer visualizations for \textsc{CrossEp-Emb} and \textsc{CrossEp-Tool}.

\subsubsection{\textsc{CrossEp-Emb}}
\label{app:transfer_embodied}

\begin{figure}[htbp]
    \centering
    \includegraphics[width=1\linewidth]{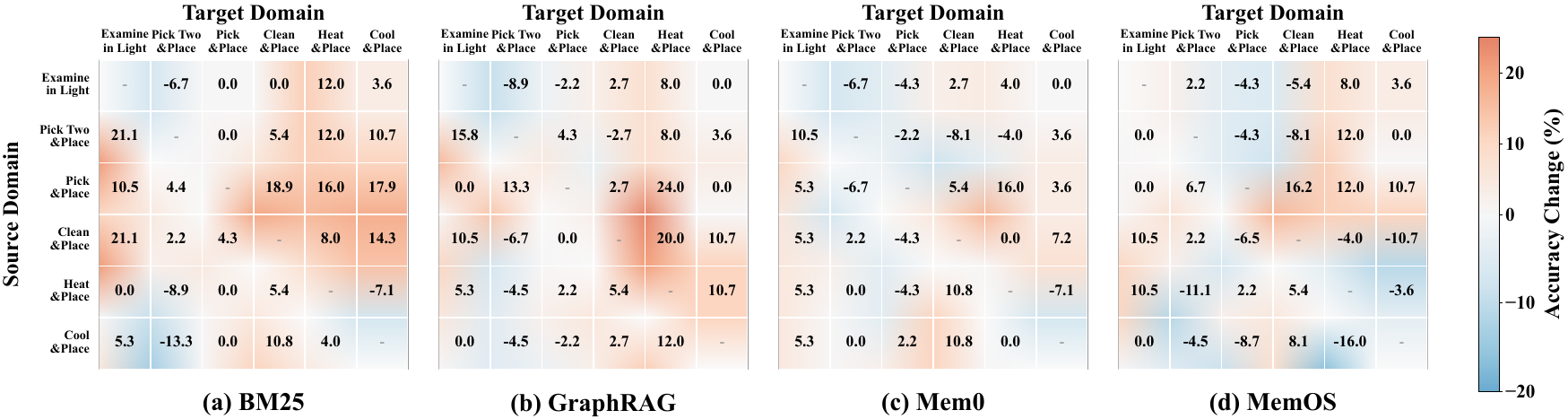}
    \caption{Cross-environment transfer results in \textsc{CrossEp-Emb} for BM25, GraphRAG, Mem0, and MemOS.}
    \label{fig:embodied_BM25_GraphRAG_Mem0_MemOS}
\end{figure}

\begin{figure}[htbp]
    \centering
    \includegraphics[width=1\linewidth]{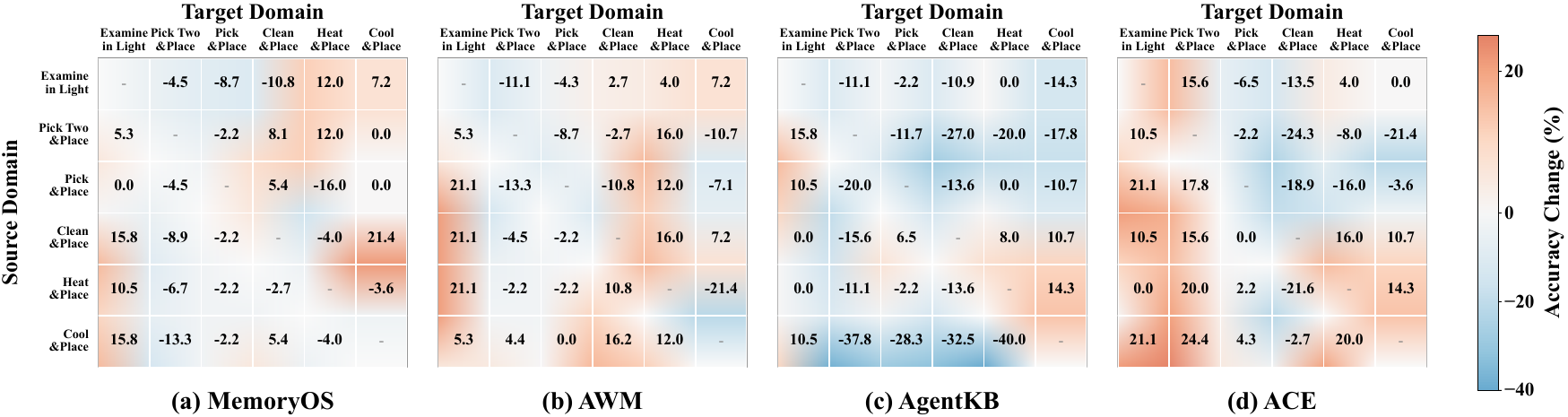}
    \caption{Cross-environment transfer results in \textsc{CrossEp-Emb} for MemoryOS, AWM, AgentKB, and ACE.}
    \label{fig:embodied_MemoryOS_AWM_AgentKB_ACE}
\end{figure}

\newpage
\subsubsection{\textsc{CrossEp-Tool}}
\label{app:transfer_tool}

\begin{figure}[htbp]
    \centering
    \includegraphics[width=1\linewidth]{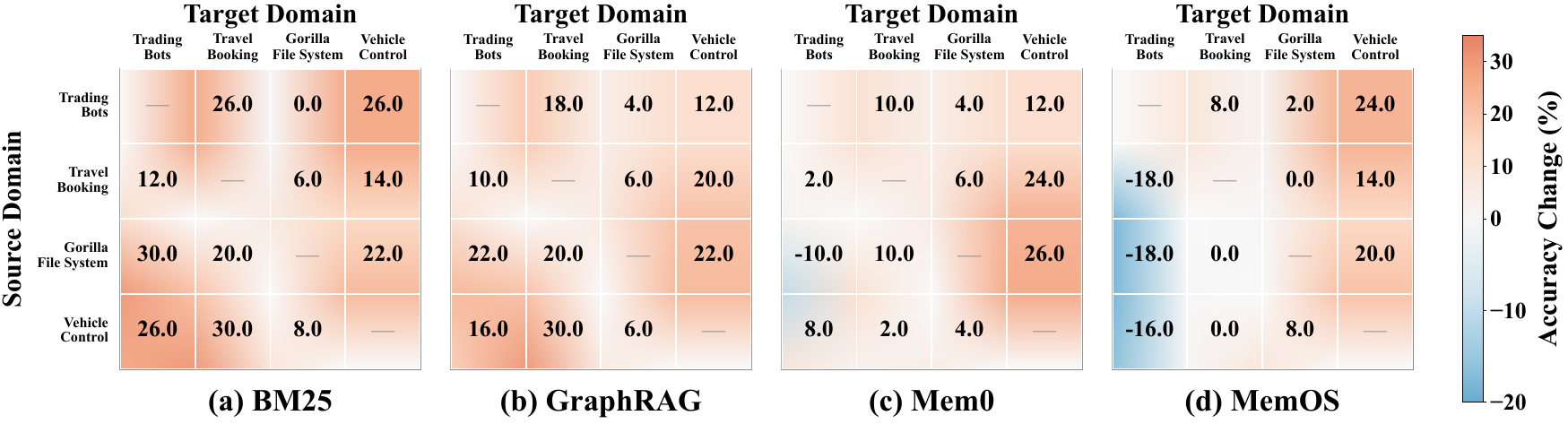}
    \caption{Cross-environment transfer results in \textsc{CrossEp-Tool} for BM25, GraphRAG, Mem0, and MemOS.}
    \label{fig:tool_BM25_GraphRAG_Mem0_MemOS}
\end{figure}

\begin{figure}[htbp]
    \centering
    \includegraphics[width=1\linewidth]{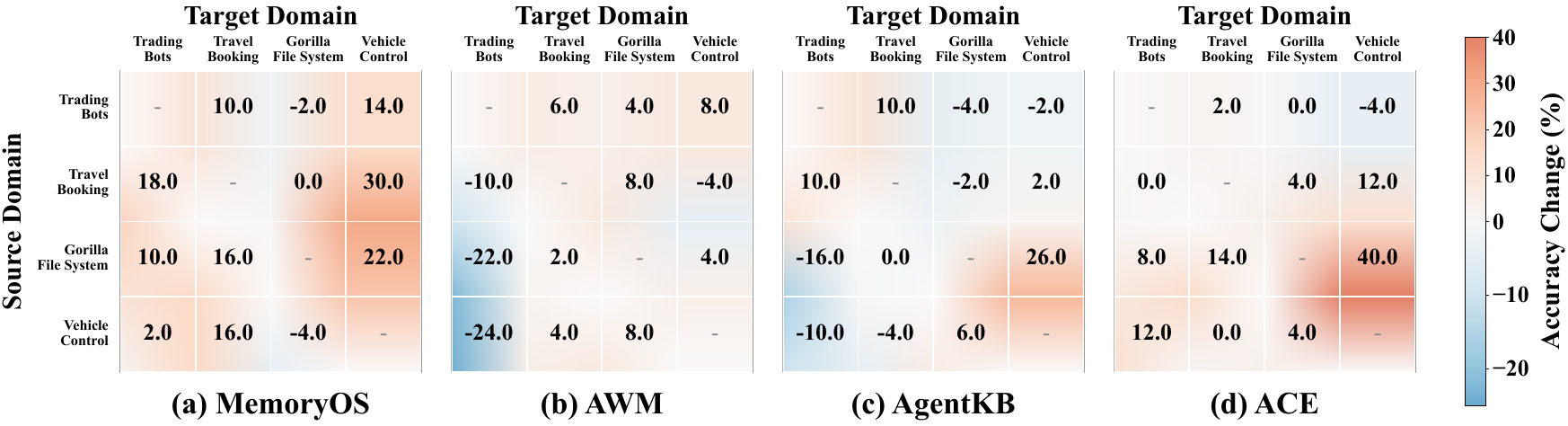}
    \caption{Cross-environment transfer results in \textsc{CrossEp-Tool} for MemoryOS, AWM, AgentKB, and ACE.}
    \label{fig:tool_MemoryOS_AWM_AgentKB_ACE}
\end{figure}

\subsection{Additional Efficiency Results for Cross-Episode Settings}
\label{app:crossep_efficiency_results}

We report token usage for cross-episode knowledge evolution and both average steps and token usage for cross-episode execution settings.

\label{app:crossep_know_efficiency}

\begin{table*}[htbp]
\centering
\caption{Efficiency comparison on \textsc{CrossEp-Know}, measured by token usage. Lower is better. The best result in each column is bolded and the runner-up is underlined.}
\label{tab:clbench_token}
\setlength{\tabcolsep}{5pt}
\renewcommand{\arraystretch}{1}
\resizebox{\textwidth}{!}{
\begin{tabular}{lcccc}
\toprule
Method
& Domain Knowledge
& Empirical Discovery
& Procedural Task
& Rule System Application \\
\midrule
Gemini-3-Flash & 15465 & 18546 & 14701 & 14942 \\
GPT-5-mini     & 17188 & 20278 & 16636 & 15295 \\
DeepSeek-V3.2  & \textbf{6416} & \textbf{8145} & \textbf{6021} & \textbf{5613} \\
\midrule
BM25           & 14610 & 16224 & 14351 & 14113 \\
Qwen3-Emb-4B   & 14576 & 16655 & 14609 & 14056 \\
GraphRAG       & 15850 & 17993 & 15773 & 15386 \\
\midrule
Mem0           & 22852 & 26336 & 21932 & 21320 \\
A-MEM          & 31939 & 35668 & 31854 & 31207 \\
MemOS          & \underline{7640} & \underline{9456} & \underline{7180} & \underline{6910} \\
MemoryOS       & 18506 & 23676 & 20074 & 19409 \\
\midrule
AWM            & 14604 & 18092 & 13776 & 13054 \\
SkillWeaver    & 14407 & 18028 & 13708 & 12683 \\
AgentKB        & 10002 & 11793 & 9532 & 9241 \\
ACE            & 20802 & 23612 & 20436 & 19495 \\
ReasoningBank  & 13632 & 15440 & 13329 & 13278 \\
\midrule
MemEvolve      & 9876 & 11734 & 9381 & 9080 \\
\bottomrule
\end{tabular}
}
\end{table*}

\label{app:crossep_emb_efficiency}

\begin{table*}[htbp]
\centering
\caption{Efficiency comparison on \textsc{CrossEp-Emb} tasks measured by average steps. Lower is better. The best result in each column is bolded and the runner-up is underlined.}
\label{tab:alfworld_steps}
\setlength{\tabcolsep}{4pt}
\renewcommand{\arraystretch}{1}
\resizebox{\textwidth}{!}{
\begin{tabular}{lccccccc}
\toprule
Method
& Examine in Light
& Pick \& Place
& Clean \& Place
& Cool \& Place
& Heat \& Place
& Pick Two \& Place
& Avg. (Weighted) \\
\midrule
Gemini-3-Flash      & 15.00 & 13.20 & 16.55 & 15.43 & 15.60 & 18.76 & 15.85 \\
GPT-5-mini          & 14.74 & 11.85 & 18.70 & 18.11 & 16.16 & 18.36 & 16.27 \\
DeepSeek-V3.2       & 11.16 & 9.22  & 15.30 & 15.71 & 16.32 & 17.16 & 14.11 \\
\midrule
BM25                & 11.26 & 8.89  & 14.49 & 14.68 & 14.08 & 15.51 & 13.10 \\
Qwen3-Emb-4B  & 12.37 & \textbf{8.57} & 13.27 & 14.68 & 14.84 & 14.84 & 12.89 \\
GraphRAG            & 12.63 & \underline{8.52} & 13.32 & 14.46 & 13.44 & 14.42 & 12.57 \\
\midrule
AgentKB             & 12.21 & 11.37 & 12.05 & 18.64 & 12.16 & 13.76 & 13.23 \\
ACE                 & 12.79 & 8.98  & 15.76 & \textbf{10.68} & \textbf{11.40} & 14.56 & 12.39 \\
ReasoningBank       & 10.11 & 10.13 & \textbf{12.00} & 13.39 & 14.44 & 14.31 & 12.41 \\
Mem0                & 11.05 & 9.41  & 13.50 & 14.23 & 14.33 & 16.78 & 13.27 \\
AWM                 & \underline{9.58} & 9.98  & 12.19 & \underline{12.32} & 11.92 & 15.82 & \underline{12.23} \\
SkillWeaver         & 12.84 & 10.37 & 15.00 & 15.50 & 13.48 & 16.04 & 13.85 \\
MemEvolve           & 12.89 & 10.00 & 14.35 & 13.79 & 12.04 & \underline{13.53} & 12.66 \\
MemOS               & 12.53 & 9.96  & 13.68 & 14.89 & 15.36 & 17.02 & 13.85 \\
MemoryOS            & 11.00 & 8.61  & 13.32 & 14.29 & 14.80 & 16.22 & 12.99 \\
A-MEM               & \textbf{9.37} & 8.89  & \underline{12.08} & 14.14 & \underline{11.60} & \textbf{12.58} & \textbf{11.43} \\
\bottomrule
\end{tabular}
}
\end{table*}

\begin{table*}[htbp]
\centering
\caption{Efficiency comparison on \textsc{CrossEp-Emb} tasks measured by token usage. Lower is better. The best result in each column is bolded and the runner-up is underlined.}
\label{tab:alfworld_token}
\setlength{\tabcolsep}{4pt}
\renewcommand{\arraystretch}{1}
\resizebox{\textwidth}{!}{
\begin{tabular}{lccccccc}
\toprule
Method
& Examine in Light
& Pick \& Place
& Clean \& Place
& Cool \& Place
& Heat \& Place
& Pick Two \& Place
& Avg. (Weighted) \\
\midrule
GPT-5-mini          & 40382 & \underline{26094} & 63944  & 67886  & 56080  & 43831  & 48043 \\
Gemini-3-Flash      & 30384 & 26783 & 61690  & 61724  & 51490  & 40447  & 44637 \\
DeepSeek-V3.2       & \textbf{25653} & \textbf{16001} & \textbf{46105} & \textbf{51950} & 53153 & \underline{38274} & \textbf{37175} \\
\midrule
BM25                & 31771 & 65227 & 118241 & 103920 & 80766  & 126938 & 93100 \\
Qwen3-Emb-4B  & 32634 & 52669 & 82326  & 86253  & 78910  & 105429 & 76105 \\
GraphRAG            & 37746 & 57691 & 100512 & 92602  & 83649  & 108669 & 83320 \\
\midrule
AgentKB             & 37423 & 33358 & 45436 & 75351  & \underline{48484} & 39087  & 45037 \\
ACE                 & 60520 & 43583 & 82866  & 59019  & 57865  & 58046  & 59660 \\
ReasoningBank       & 33244 & 29045 & \underline{45198} & 57985  & 61394  & 42740  & 43609 \\
Mem0                & 40418 & 31188 & 50676  & 59465  & 59930  & 53237  & 48182 \\
AWM                 & \underline{28796} & 31693 & 51074  & 58498 & 55304  & 50474  & 45933 \\
SkillWeaver         & 45981 & 34945 & 62774  & 71643  & 62964  & 49480  & 53052 \\
MemEvolve           & 38887 & 26915 & 49619  & \underline{52045}  & \textbf{41926} & \textbf{35635} & \underline{39609} \\
MemOS               & 39318 & 26133 & 51165  & 65274  & 60703  & 49216  & 47011 \\
MemoryOS            & 77670 & 55638 & 112278 & 134858 & 136843 & 102816 & 100066 \\
A-MEM               & 53004 & 37768 & 67852  & 73298  & 60338  & 58932  & 57338 \\
\bottomrule
\end{tabular}
}
\end{table*}

\label{app:crossep_tool_efficiency}

\begin{table*}[htbp]
\centering
\caption{Efficiency comparison on \textsc{CrossEp-Tool}, measured by average steps. Lower is better. The best result in each column is bolded and the runner-up is underlined.}
\label{tab:bfcl_ce_steps}
\setlength{\tabcolsep}{5pt}
\renewcommand{\arraystretch}{1}

\begin{tabular}{lccccc}
\toprule
Method
& \begin{tabular}[c]{@{}c@{}}Gorilla File\\System\end{tabular}
& \begin{tabular}[c]{@{}c@{}}Vehicle\\Control\end{tabular}
& \begin{tabular}[c]{@{}c@{}}Trading\\Bots\end{tabular}
& \begin{tabular}[c]{@{}c@{}}Travel\\Booking\end{tabular}
& Average \\
\midrule
Gemini-3-Flash   & \underline{12.40} & 14.08 & 12.28 & 12.86 & 12.90 \\
GPT-5-mini       & \textbf{12.18}    & \textbf{12.70} & \underline{10.40} & 10.68 & \textbf{11.49} \\
DeepSeek-V3.2    & 14.82             & \underline{13.44} & \underline{10.40} & 10.90 & \underline{12.39} \\
\midrule
BM25             & 18.08             & 15.94 & 12.52 & 11.80 & 14.59 \\
Qwen3-Emb-4B & 17.50           & 15.54 & 12.16 & 10.82 & 14.01 \\
GraphRAG         & 19.20             & 14.28 & 11.78 & 11.00 & 14.07 \\
\midrule
ReasoningBank    & 18.92             & 15.66 & 12.16 & 12.16 & 14.73 \\
ACE              & 18.62             & 16.02 & \textbf{7.94} & \underline{10.60} & 13.30 \\
Mem0             & 15.00             & 14.94 & 10.86 & 11.42 & 13.06 \\
AgentKB          & 23.44             & 20.16 & 13.26 & 12.50 & 17.33 \\
AWM              & 17.70             & 14.48 & 12.00 & \textbf{10.42} & 13.65 \\
SkillWeaver      & 16.88             & 13.96 & 10.88 & 11.36 & 13.27 \\
MemEvolve        & 17.90             & 15.54 & 11.18 & 11.32 & 13.99 \\
MemOS            & 15.76             & 14.30 & 10.32 & 11.40 & 12.95 \\
MemoryOS         & 19.38             & 15.02 & 15.82 & 12.30 & 15.63 \\
A-MEM            & 19.48             & 13.98 & 12.56 & 10.76 & 13.20 \\
\bottomrule
\end{tabular}

\end{table*}

\begin{table*}[htbp]
\centering
\caption{Efficiency comparison on \textsc{CrossEp-Tool}, measured by token usage. Lower is better. The best result in each column is bolded and the runner-up is underlined.}
\label{tab:bfcl_ce_tokens}
\setlength{\tabcolsep}{5pt}
\renewcommand{\arraystretch}{1}

\begin{tabular}{lccccc}
\toprule
Method
& \begin{tabular}[c]{@{}c@{}}Gorilla File\\System\end{tabular}
& \begin{tabular}[c]{@{}c@{}}Vehicle\\Control\end{tabular}
& \begin{tabular}[c]{@{}c@{}}Trading\\Bots\end{tabular}
& \begin{tabular}[c]{@{}c@{}}Travel\\Booking\end{tabular}
& Average \\
\midrule
Gemini-3-Flash   & \underline{46221} & \underline{51252} & \underline{37416} & 57043 & \underline{47983} \\
GPT-5-mini       & \textbf{36699}    & \textbf{39325}    & \textbf{26520} & \textbf{37108} & \textbf{34913} \\
DeepSeek-V3.2    & 67664             & 60192             & 39266 & \underline{56026} & 55787 \\
\midrule
BM25             & 254111            & 221727            & 167244 & 168780 & 202966 \\
Qwen3-Emb-4B & 230714          & 203511            & 167175 & 144507 & 186477 \\
GraphRAG         & 270176            & 193998            & 158727 & 154069 & 194243 \\
\midrule
ReasoningBank    & 102835            & 83237             & 59223 & 74413 & 79927 \\
ACE              & 139608            & 127034            & 67865 & 102581 & 109272 \\
Mem0             & 79390             & 79831             & 54195 & 70791 & 71052 \\
AgentKB          & 145361            & 134314            & 78242 & 91854 & 112333 \\
AWM              & 92055             & 74334             & 58037 & 60606 & 71258 \\
SkillWeaver      & 83388             & 67458 & 47794 & 63644 & 65571 \\
MemEvolve        & 97691             & 85105             & 54285 & 71952 & 77258 \\
MemOS            & 79324             & 70373             & 44716 & 65292 & 64926 \\
MemoryOS         & 125445            & 102908            & 181489 & 102188 & 128033 \\
A-MEM            & 217512            & 153201            & 210504 & 143973 & 181298 \\
\bottomrule
\end{tabular}

\end{table*}

\label{app:crossep_web_efficiency}

\begin{table*}[htbp]
\centering
\caption{Efficiency comparison on \textsc{CrossEp-Web}, measured by average steps. Lower is better. The best result in each column is bolded and the runner-up is underlined.}
\label{tab:websearch_steps}
\setlength{\tabcolsep}{5pt}
\renewcommand{\arraystretch}{1}

\begin{tabular}{lcccc}
\toprule
Method
& xbench
& WebWalkerQA
& xbench$\rightarrow$WebWalkerQA
& WebWalkerQA$\rightarrow$xbench \\
\midrule
Gemini-3-Flash   & 11.10 & 9.95 & - & - \\
GPT-5-mini       & 12.63             & 10.07 & - & - \\
DeepSeek-V3.2    & \textbf{10.01}    & \textbf{8.12} & - & - \\
\midrule
BM25             & \underline{10.92}             & 10.24 & 10.44 & 12.06 \\
Qwen3-Emb-4B & 12.41           & 9.82 & 9.84 & 11.65 \\
GraphRAG         & 11.68             & \underline{8.81} & 10.55 & 11.40 \\
\midrule
ReasoningBank    & 11.92             & 9.59 & \textbf{9.29} & 11.39 \\
ACE              & 12.92             & 10.09 & 10.48 & \textbf{10.59} \\
Mem0             & 12.20             & 9.74 & 10.01 & 12.79 \\
AgentKB          & 12.62             & 9.85 & 10.46 & 11.49 \\
AWM              & 12.19             & 9.51 & 10.09 & \underline{10.94} \\
SkillWeaver      & 12.08             & 9.62 & 9.96 & 11.79 \\
MemEvolve        & 12.31             & 9.84 & \underline{9.81} & 11.65 \\
MemOS            & 11.17             & 9.84 & 10.17 &  14.49 \\
MemoryOS         & 14.24             & 12.08 & 12.55 & 13.89 \\
A-MEM            & 14.20             & 11.31 & 12.35 & 12.65 \\
\bottomrule
\end{tabular}

\end{table*}

\begin{table*}[htbp]
\centering
\caption{Efficiency comparison on \textsc{CrossEp-Web}, measured by token usage. Lower is better. The best result in each column is bolded and the runner-up is underlined.}
\label{tab:websearch_tokens}
\setlength{\tabcolsep}{5pt}
\renewcommand{\arraystretch}{1}

\begin{tabular}{lcccc}
\toprule
Method
& xbench
& WebWalkerQA
& xbench$\rightarrow$WebWalkerQA
& WebWalkerQA$\rightarrow$xbench \\
\midrule
Gemini-3-Flash   & \underline{269911} & 140189 & - & - \\
GPT-5-mini       & 306538             & 138425 & - & - \\
DeepSeek-V3.2    & 377624             & 164081 & - & - \\
\midrule
BM25             & 384243             & 180381 & 302560 & 431878 \\
Qwen3-Emb-4B & 451976           & 262780 & 256687 & 412250 \\
GraphRAG         & 432288             & 220649 & 289542 & 402795 \\
\midrule
ReasoningBank    & 287683             & \underline{132898} & \textbf{131799} & 281394 \\
ACE              & 481140             & 366907 & 409670 & 815386 \\
Mem0             & 316996             & 138051 & \underline{137435} & 317055 \\
AgentKB          & 344953             & 136622 & 156246 & \underline{266775} \\
AWM              & 301893             & 138651 & 150348 & \textbf{259937} \\
SkillWeaver      & 293638 & \textbf{130577} & 143156 & 286286 \\
MemEvolve        & 294272             & 142265 & 145114 & 284052 \\
MemOS            & \textbf{260389}    & 140872 & 156330 & 316401 \\
MemoryOS         & 280934             & 161801 & 151790 & 309810 \\
A-MEM            & 553566             & 340888 & 396823 & 454250 \\
\bottomrule
\end{tabular}
\end{table*}

\newpage

\subsection{Additional Results for \textsc{InEp-Exec}}
\label{app:inep_exec_additional}

We report complementary results for \textsc{InEp-Exec}, including progress score and average steps under different context budgets.

\begin{table*}[htbp]
\centering
\caption{Progress score (\%) comparison across different memory methods on \textsc{InEp-Exec} under different context lengths. Unless otherwise specified, all memory agents use DeepSeek-V3.2 as the backbone. Best results are bolded, and runner-up results among memory methods are underlined.}
\label{tab:memory_benchmark_ps}
\setlength{\tabcolsep}{3pt}
\renewcommand{\arraystretch}{1.15}
\resizebox{\textwidth}{!}{
\begin{threeparttable}
\begin{tabular}{l|cccc|cccc|cccc|cccc|cccc}
\toprule
\multirow{2}{*}{Method} 
& \multicolumn{4}{c|}{Gorilla File System}
& \multicolumn{4}{c|}{Vehicle Control}
& \multicolumn{4}{c|}{Trading Bots}
& \multicolumn{4}{c|}{Travel Booking}
& \multicolumn{4}{c}{Overall} \\
\cmidrule(lr){2-5}
\cmidrule(lr){6-9}
\cmidrule(lr){10-13}
\cmidrule(lr){14-17}
\cmidrule(lr){18-21}
& 16K & 32K & 64K & 128K
& 16K & 32K & 64K & 128K
& 16K & 32K & 64K & 128K
& 16K & 32K & 64K & 128K
& 16K & 32K & 64K & 128K \\
\midrule

\multicolumn{21}{c}{\emph{Long-Context LLM}} \\[2pt]
Gemini-3-Flash 
& \textbf{77.4} & 74.0 & \textbf{77.6} & 75.2
& \textbf{76.9} & \textbf{80.6} & \textbf{77.2} & \textbf{71.2}
& \textbf{52.5} & \textbf{91.2} & \textbf{85.2} & \textbf{88.7}
& \textbf{76.2} & \textbf{78.5} & \textbf{75.2} & \textbf{70.1}
& \textbf{70.7} & \textbf{81.0} & \textbf{78.8} & \textbf{76.3} \\
GPT-5-mini 
& 58.7 & 61.8 & 63.7 & 62.4
& 62.3 & 59.9 & 56.2 & 55.9
& 48.5 & 78.7 & 82.0 & 80.6
& 56.7 & 53.2 & 57.7 & 54.0
& 56.5 & 63.4 & 64.9 & 63.2 \\
\rowcolor{headergray}
DeepSeek-V3.2 
& 76.9 & \textbf{75.5} & 73.2 & \textbf{76.2}
& 57.2 & 60.1 & 60.2 & 60.5
& 29.4 & 50.3 & 54.6 & 52.2
& 40.2 & 37.9 & 35.8 & 40.3
& 50.9 & 56.0 & 55.9 & 57.3 \\

\specialrule{0.4pt}{1pt}{2pt}
\specialrule{0.4pt}{0pt}{3pt}

\multicolumn{21}{c}{\emph{Retrieval-Augmented Memory}} \\[2pt]
BM25 
& 77.4 & \underline{80.9} & 72.8 & 76.5
& \underline{68.8} & 63.6 & 58.3 & 65.0
& 34.1 & 63.2 & 54.5 & 48.7
& 47.6 & 47.4 & 38.8 & 45.0
& 57.0 & \underline{63.8} & 56.1 & 58.8 \\
Qwen3-Emb-4B 
& 58.7 & \textbf{81.1} & 68.7 & \underline{80.3}
& 65.0 & 64.7 & 60.6 & 58.4
& 37.7 & 47.6 & 55.6 & 50.0
& \textbf{54.7} & 45.3 & 49.1 & 46.3
& 54.0 & 59.7 & 58.5 & 58.8 \\
GraphRAG 
& 76.9 & 80.2 & \textbf{80.2} & 79.1
& 61.5 & 58.8 & 64.4 & 60.6
& 33.8 & 43.4 & 56.8 & 47.5
& 45.0 & 48.4 & \textbf{50.8} & 45.4
& 54.3 & 57.7 & 63.0 & 58.2 \\

\addlinespace[2pt]
\multicolumn{21}{c}{\emph{Short-Term Memory}} \\[2pt]
MemAgent 
& 70.6 & 74.8 & \underline{79.8} & 72.0
& 56.5 & 51.3 & 52.4 & 56.4
& 28.8 & 49.8 & 52.9 & \underline{60.0}
& 28.0 & 43.6 & 36.2 & 37.3
& 46.0 & 54.9 & 55.3 & 56.4 \\
MemoBrain 
& 69.0 & 72.5 & 78.6 & 73.4
& 60.7 & 54.5 & 60.3 & 61.9
& 24.8 & 44.9 & 57.0 & 50.4
& 42.1 & 35.4 & 40.7 & 44.9
& 49.1 & 51.8 & 59.1 & 57.6 \\

\addlinespace[2pt]
\multicolumn{21}{c}{\emph{General Long-Term Memory}} \\[2pt]
Mem0 
& \underline{78.7} & 70.4 & 70.9 & \textbf{82.5}
& 63.9 & 62.0 & 59.8 & 58.5
& 36.8 & 56.6 & 47.0 & 46.4
& 46.2 & 41.0 & 43.1 & 47.2
& 56.4 & 57.5 & 55.2 & 58.6 \\
A-MEM 
& 76.4 & 75.0 & 76.2 & 72.8
& 67.2 & 62.9 & 54.8 & 65.7
& 37.2 & 43.8 & 56.4 & 47.2
& 41.3 & 48.4 & 44.7 & 43.3
& 55.5 & 57.5 & 58.0 & 57.3 \\
MemOS 
& 76.9 & 75.2 & 79.0 & 76.0
& 63.6 & 55.4 & 61.0 & 57.2
& 32.8 & 51.9 & 55.2 & 53.9
& 45.7 & 34.4 & 37.0 & 35.3
& 54.8 & 54.2 & 58.0 & 55.6 \\
MemoryOS 
& 75.4 & 79.9 & 72.5 & 73.6
& 57.8 & 54.9 & 62.6 & 61.2
& 26.4 & 53.6 & 50.0 & 21.5
& 38.7 & 42.1 & 39.5 & 49.8
& 49.6 & 57.6 & 56.1 & 51.5 \\

\addlinespace[2pt]
\multicolumn{21}{c}{\emph{Procedural Long-Term Memory}} \\[2pt]
AWM 
& 75.2 & 74.9 & 78.3 & 77.9
& 60.0 & 65.8 & 63.8 & 50.2
& 37.0 & \textbf{70.7} & \textbf{85.0} & 52.3
& 43.9 & \underline{50.3} & 47.0 & 42.4
& 54.0 & \textbf{65.4} & \textbf{68.5} & 55.7 \\
SkillWeaver 
& 75.1 & 72.6 & 73.8 & 74.9
& 61.3 & 63.0 & \underline{69.4} & 67.4
& 31.4 & 49.6 & 67.7 & 52.3
& 46.4 & 47.0 & \underline{50.7} & \underline{51.5}
& 53.5 & 58.0 & 65.4 & 61.5 \\
AgentKB 
& 75.0 & 79.9 & 77.2 & 74.0
& 22.6 & 61.6 & 58.1 & \textbf{74.7}
& \textbf{46.2} & 48.3 & \underline{75.6} & \textbf{66.6}
& 44.7 & 39.1 & 36.1 & 39.9
& 47.1 & 57.2 & 61.7 & 63.8 \\
ACE 
& 76.6 & 70.8 & 78.4 & 78.1
& 59.3 & 51.8 & 53.7 & 59.9
& 26.8 & 48.0 & 52.9 & 52.6
& 45.3 & 40.9 & 43.7 & 41.6
& 52.0 & 52.8 & 57.2 & 58.0 \\
ReasoningBank 
& \textbf{83.3} & 75.4 & 74.6 & 79.0
& \textbf{69.6} & \underline{65.9} & 64.7 & 67.5
& 32.3 & 50.9 & 45.0 & 56.7
& 50.4 & \textbf{60.8} & 45.2 & \textbf{53.6}
& \underline{58.9} & 63.2 & 57.4 & \underline{64.2} \\

\addlinespace[2pt]
\multicolumn{21}{c}{\emph{Meta-Evolution Memory}} \\[2pt]
MemEvolve 
& 74.1 & 73.5 & 74.8 & 79.1
& 67.7 & \textbf{66.1} & \textbf{79.1} & \underline{69.0}
& \underline{41.0} & \underline{64.2} & 75.0 & 58.3
& \underline{53.9} & 45.1 & 44.2 & 50.8
& \textbf{59.2} & 62.2 & \underline{68.3} & \textbf{64.3} \\

\bottomrule
\end{tabular}
\end{threeparttable}
}
\end{table*}

\begin{table*}[htbp]
\centering
\caption{Efficiency comparison on \textsc{InEp-Exec} under the 16K setting, measured by average steps. Lower is better. The best result in each column is bolded and the runner-up is underlined.}
\label{tab:bfcl_ie_longcontext_efficiency_16k}
\setlength{\tabcolsep}{5pt}
\renewcommand{\arraystretch}{0.8}

\begin{tabular}{lccccc}
\toprule
Method
& \begin{tabular}[c]{@{}c@{}}Gorilla File\\System\end{tabular}
& \begin{tabular}[c]{@{}c@{}}Vehicle\\Control\end{tabular}
& \begin{tabular}[c]{@{}c@{}}Trading\\Bots\end{tabular}
& \begin{tabular}[c]{@{}c@{}}Travel\\Booking\end{tabular}
& Average \\
\midrule
Gemini-3-Flash & \underline{12.73} & 13.73 & 17.02 & 13.22 & 14.11 \\
GPT-5-mini     & \textbf{12.56}    & \underline{12.72} & 17.06 & \textbf{10.16} & 13.12 \\
DeepSeek-V3.2  & 15.58             & 13.92 & \textbf{10.12} & 10.74 & \underline{12.59} \\
\midrule
BM25           & 16.40             & 14.64 & 12.28 & 11.00 & 13.58 \\
Qwen3-Emb-4B   & 16.32             & 15.30 & 16.32 & 11.12 & 14.77 \\
GraphRAG       & 16.46             & 14.40 & 11.24 & 11.00 & 13.28 \\
\midrule
MemAgent       & 15.48             & 14.20 & \underline{10.16} & \underline{10.26} & \textbf{12.53} \\
MemoBrain      & 15.66             & 13.83 & 12.35 & 10.88 & 13.18 \\
\midrule
Mem0           & 16.40             & 14.10 & 12.32 & 11.06 & 12.99 \\
A-MEM          & 15.18             & 14.02 & 12.08 & 10.58 & 12.96 \\
MemOS          & 15.60             & 14.22 & 11.96 & 10.70 & 13.12 \\
MemoryOS       & 16.14             & 14.62 & 11.36 & 10.48 & 13.15 \\
\midrule
AWM            & 20.40             & 14.52 & 12.90 & 10.98 & 14.70 \\
SkillWeaver    & 16.88             & 14.02 & 10.42 & 11.00 & 13.08 \\
AgentKB        & 30.58             & \textbf{12.32} & 18.20 & 12.32 & 18.36 \\
ACE            & 17.20             & 14.82 & 12.72 & 11.92 & 14.16 \\
ReasoningBank  & 19.76             & 15.56 & 13.86 & 12.70 & 15.47 \\
\midrule
MemEvolve      & 17.86             & 14.76 & 11.46 & 11.32 & 13.85 \\
\bottomrule
\end{tabular}

\end{table*}

\begin{table*}[htbp]
\centering
\caption{Efficiency comparison on \textsc{InEp-Exec} under the 16K setting, measured by token usage. Lower is better. The best result in each column is bolded and the runner-up is underlined.}
\label{tab:bfcl_ie_longcontext_token_16k}
\setlength{\tabcolsep}{5pt}
\renewcommand{\arraystretch}{0.8}

\begin{tabular}{lccccc}
\toprule
Method
& \begin{tabular}[c]{@{}c@{}}Gorilla File\\System\end{tabular}
& \begin{tabular}[c]{@{}c@{}}Vehicle\\Control\end{tabular}
& \begin{tabular}[c]{@{}c@{}}Trading\\Bots\end{tabular}
& \begin{tabular}[c]{@{}c@{}}Travel\\Booking\end{tabular}
& Average \\
\midrule
Gemini-3-Flash & \underline{57428} & \underline{86939}             & 119487            & 77563             & 84594 \\
GPT-5-mini     & \textbf{46265}    & \textbf{66404}    & 129870            & \textbf{46859}    & \underline{72349} \\
DeepSeek-V3.2  & 82104             & 97262             & \underline{54125} & 65028 & 74630 \\
\midrule
BM25           & 122644            & 123672            & 101003            & 89700             & 109255 \\
Qwen3-Emb-4B   & 121698            & 127476            & 261780            & 91906             & 150715 \\
GraphRAG       & 127673            & 125803            & 87847             & 97989             & 109828 \\
\midrule
MemAgent       & 82054             & 96510 & \textbf{45461}    & \underline{61729}             & \textbf{71438} \\
MemoBrain      & 126606            & 141363            & 155852            & 89061             & 127947 \\
\midrule
Mem0           & 133415            & 153265            & 116864            & 119538            & 130771 \\
A-MEM          & 105199            & 116426            & 97830             & 92583             & 103009 \\
MemOS          & 101334            & 121539            & 114826            & 83931             & 105407 \\
MemoryOS       & 116687            & 119788            & 70983             & 85381             & 98210 \\
\midrule
AWM            & 140220            & 115285            & 81033             & 79209             & 103937 \\
SkillWeaver    & 95428             & 97980             & 59924             & 71618             & 81237 \\
AgentKB        & 227756            & 99861             & 115654            & 87945             & 132804 \\
ACE            & 143073            & 165384            & 154381            & 128670            & 147877 \\
ReasoningBank  & 153539            & 157398            & 176499            & 117945            & 151345 \\
\midrule
MemEvolve      & 106880            & 112956            & 73194             & 76621             & 92412 \\
\bottomrule
\end{tabular}

\end{table*}

\begin{table*}[htbp]
\centering
\caption{Efficiency comparison on \textsc{InEp-Exec} under the 32K setting, measured by average steps. Lower is better. The best result in each column is bolded and the runner-up is underlined.}
\label{tab:bfcl_ie_longcontext_efficiency_32k}
\setlength{\tabcolsep}{5pt}
\renewcommand{\arraystretch}{0.8}

\begin{tabular}{lccccc}
\toprule
Method
& \begin{tabular}[c]{@{}c@{}}Gorilla File\\System\end{tabular}
& \begin{tabular}[c]{@{}c@{}}Vehicle\\Control\end{tabular}
& \begin{tabular}[c]{@{}c@{}}Trading\\Bots\end{tabular}
& \begin{tabular}[c]{@{}c@{}}Travel\\Booking\end{tabular}
& Average \\
\midrule
Gemini-3-Flash & \underline{13.17} & 13.80 & 12.46 & 13.54 & 13.24 \\
GPT-5-mini     & \textbf{12.40}    & \textbf{12.80} & 10.63 & 10.52 & \textbf{11.59} \\
DeepSeek-V3.2  & 15.60             & \underline{13.36} & 11.36 & \textbf{10.04} & 12.59 \\
\midrule
BM25           & 16.56             & 14.40 & 11.28 & 11.04 & 13.32 \\
Qwen3-Emb-4B   & 15.90             & 14.44 & 12.36 & 11.36 & 13.52 \\
GraphRAG       & 15.62             & 14.46 & 11.60 & 11.34 & 13.26 \\
\midrule
MemAgent       & 15.62             & 14.10 & 10.56 & 10.80 & 12.77 \\
MemoBrain      & 15.30             & 13.54 & \underline{10.32} & \underline{10.28} & \underline{12.36} \\
\midrule
Mem0           & 16.14             & 14.20 & 10.78 & 10.84 & 12.99 \\
A-MEM          & 16.66             & 14.40 & 10.78 & 10.70 & 13.13 \\
MemOS          & 16.68             & 14.28 & 10.62 & 10.32 & 12.97 \\
MemoryOS       & 16.00             & 14.26 & 11.68 & 11.20 & 13.29 \\
\midrule
AWM            & 22.52             & 14.66 & 12.90 & 11.18 & 15.32 \\
SkillWeaver    & 16.62             & 14.44 & \textbf{10.28} & 10.58 & 12.98 \\
AgentKB        & 27.72             & 17.66 & 17.28 & 11.58 & 18.56 \\
ACE            & 18.58             & 14.96 & 11.70 & 11.64 & 14.22 \\
ReasoningBank  & 20.38             & 15.76 & 11.34 & 12.46 & 14.98 \\
\midrule
MemEvolve      & 18.02             & 14.42 & 11.28 & 11.24 & 13.74 \\
\bottomrule
\end{tabular}

\end{table*}

\begin{table*}[htbp]
\centering
\caption{Efficiency comparison on \textsc{InEp-Exec} under the 32K setting, measured by token usage. Lower is better. The best result in each column is bolded and the runner-up is underlined.}
\label{tab:bfcl_ie_longcontext_token_32k}
\setlength{\tabcolsep}{5pt}
\renewcommand{\arraystretch}{0.8}

\begin{tabular}{lccccc}
\toprule
Method
& \begin{tabular}[c]{@{}c@{}}Gorilla File\\System\end{tabular}
& \begin{tabular}[c]{@{}c@{}}Vehicle\\Control\end{tabular}
& \begin{tabular}[c]{@{}c@{}}Trading\\Bots\end{tabular}
& \begin{tabular}[c]{@{}c@{}}Travel\\Booking\end{tabular}
& Average \\
\midrule
Gemini-3-Flash & \underline{59242} & \underline{84186} & 240152            & 78847             & 116630 \\
GPT-5-mini     & \textbf{44416}    & \textbf{68534}    & 165160   & \textbf{48676}    & \textbf{81277} \\
DeepSeek-V3.2  & 82745             & 89755             & 184094            & \underline{58836} & 103858 \\
\midrule
BM25           & 137230            & 155033            & 166033            & 106051            & 141087 \\
Qwen3-Emb-4B   & 133863            & 172219            & 188112            & 113235            & 151857 \\
GraphRAG       & 131937            & 174512            & 168065            & 118035            & 148137 \\
\midrule
MemAgent       & 84526             & 100590            & \textbf{164319} & 64233            & \underline{103417} \\
MemoBrain      & 128844            & 127004            & 173458            & 77045           & 126588 \\
\midrule
Mem0           & 131693            & 156635            & 222465            & 117150            & 156986 \\
A-MEM          & 126692            & 135192            & 157307            & 103522            & 130678 \\
MemOS          & 109562            & 123567            & 194886            & 81028             & 127261 \\
MemoryOS       & 137656            & 187311            & 172611            & 132442            & 157430 \\
\midrule
AWM            & 179134            & 148277            & 218708            & 94064             & 160046 \\
SkillWeaver    & 95321             & 104407            & \underline{164551}            & 68708             & 108247 \\
AgentKB        & 204919            & 206348            & 280370            & 84725             & 194090 \\
ACE            & 151680            & 174171            & 258134            & 126804            & 177697 \\
ReasoningBank  & 164874            & 187292            & 243550            & 117035            & 178188 \\
\midrule
MemEvolve      & 113777            & 115031            & 188065            & 76205             & 123270 \\
\bottomrule
\end{tabular}

\end{table*}

\begin{table*}[htbp]
\centering
\caption{Efficiency comparison on \textsc{InEp-Exec} under the 64K setting, measured by steps. Lower is better. The best result in each column is bolded and the runner-up is underlined.}
\label{tab:bfcl_ie_longcontext_efficiency_64k}
\setlength{\tabcolsep}{5pt}
\renewcommand{\arraystretch}{0.8}

\begin{tabular}{lccccc}
\toprule
Method
& \begin{tabular}[c]{@{}c@{}}Gorilla File\\System\end{tabular}
& \begin{tabular}[c]{@{}c@{}}Vehicle\\Control\end{tabular}
& \begin{tabular}[c]{@{}c@{}}Trading\\Bots\end{tabular}
& \begin{tabular}[c]{@{}c@{}}Travel\\Booking\end{tabular}
& Average \\
\midrule
Gemini-3-Flash & \textbf{12.54}    & 13.87             & 12.15             & 12.88             & 12.85 \\
GPT-5-mini     & \underline{12.62} & \textbf{12.82}    & \textbf{10.38}    & \underline{10.30} & \textbf{11.53} \\
DeepSeek-V3.2  & 15.72             & 14.32             & \underline{10.48} & 10.64             & 12.79 \\
\midrule
BM25           & 16.22             & 14.46             & 10.78             & 11.24             & 13.18 \\
Qwen3-Emb-4B   & 16.04             & 14.64             & 11.02             & 11.32             & 13.26 \\
GraphRAG       & 16.40             & 14.40             & 11.12             & 11.30             & 13.30 \\
\midrule
MemAgent       & 15.68             & 13.88             & 10.70             & \textbf{10.10}    & \underline{12.59} \\
MemoBrain      & 15.78             & \underline{13.44} & 11.04             & 10.38             & 12.66 \\
\midrule
Mem0           & 16.22             & 13.92             & 10.56             & 10.50             & 12.80 \\
A-MEM          & 15.82             & 14.06             & 11.04             & 10.96             & 12.97 \\
MemOS          & 15.40             & 14.16             & \underline{10.48} & 10.44             & 12.62 \\
MemoryOS       & 16.84             & 14.68             & 10.84             & 10.98             & 13.34 \\
\midrule
AWM            & 23.00             & 15.00             & 12.68             & 10.94             & 15.41 \\
SkillWeaver    & 15.28             & 14.34             & 11.68             & 10.98             & 13.07 \\
AgentKB        & 28.82             & 18.06             & 20.14             & 12.62             & 19.91 \\
ACE            & 19.44             & 14.48             & 11.62             & 11.86             & 14.35 \\
ReasoningBank  & 21.30             & 15.22             & 11.58             & 11.88             & 14.99 \\
\midrule
MemEvolve      & 19.22             & 14.54             & 12.62             & 11.18             & 14.39 \\
\bottomrule
\end{tabular}

\end{table*}

\begin{table*}[htbp]
\centering
\caption{Efficiency comparison on \textsc{InEp-Exec} under the 64K setting, measured by token usage. Lower is better. The best result in each column is bolded and the runner-up is underlined.}
\label{tab:bfcl_ie_longcontext_token_64k}
\setlength{\tabcolsep}{5pt}
\renewcommand{\arraystretch}{0.8}

\begin{tabular}{lccccc}
\toprule
Method
& \begin{tabular}[c]{@{}c@{}}Gorilla File\\System\end{tabular}
& \begin{tabular}[c]{@{}c@{}}Vehicle\\Control\end{tabular}
& \begin{tabular}[c]{@{}c@{}}Trading\\Bots\end{tabular}
& \begin{tabular}[c]{@{}c@{}}Travel\\Booking\end{tabular}
& Average \\
\midrule
Gemini-3-Flash & \underline{55926} & \underline{84115} & 288709 & 74250 & 124675 \\
GPT-5-mini     & \textbf{45294}    & \textbf{68810}    & \textbf{178818} & \textbf{47621} & \textbf{85136} \\
DeepSeek-V3.2  & 83828             & 101644            & \underline{186160} & 63713 & \underline{108837} \\
\midrule
BM25           & 134793            & 154066            & 256800 & 108267 & 163482 \\
Qwen3-Emb-4B   & 131000            & 179759            & 300753 & 111326 & 180709 \\
GraphRAG       & 137896            & 167379            & 235670 & 109425 & 162592 \\
\midrule
MemAgent       & 85190             & 97681 & 194087 & \underline{59835} & 109198 \\
MemoBrain      & 108767            & 105150            & 233231 & 74542 & 130422 \\
\midrule
Mem0           & 134828            & 152657            & 224629 & 114879 & 156748 \\
A-MEM          & 112351            & 129972            & 256085 & 103096 & 150426 \\
MemOS          & 96967             & 123145            & 225063 & 82052 & 131807 \\
MemoryOS       & 149634            & 240353            & 250932 & 146350 & 196817 \\
\midrule
AWM            & 181790            & 163050            & 302457 & 91774 & 184768 \\
SkillWeaver    & 84013             & 105967            & 246821 & 71215 & 127004 \\
AgentKB        & 217087            & 235739            & 654378 & 93989 & 300298 \\
ACE            & 165695            & 174011            & 332247 & 128942 & 200224 \\
ReasoningBank  & 175052            & 169242            & 299487 & 108938 & 188180 \\
\midrule
MemEvolve      & 118371            & 113117            & 303296 & 77040 & 152956 \\
\bottomrule
\end{tabular}

\end{table*}

\begin{table*}[htbp]
\centering
\caption{Efficiency comparison on \textsc{InEp-Exec} under the 128K setting, measured by average steps. Lower is better. The best result in each column is bolded and the runner-up is underlined.}
\label{tab:bfcl_ie_longcontext_efficiency_128k}
\setlength{\tabcolsep}{5pt}
\renewcommand{\arraystretch}{0.8}

\begin{tabular}{lccccc}
\toprule
Method
& \begin{tabular}[c]{@{}c@{}}Gorilla File\\System\end{tabular}
& \begin{tabular}[c]{@{}c@{}}Vehicle\\Control\end{tabular}
& \begin{tabular}[c]{@{}c@{}}Trading\\Bots\end{tabular}
& \begin{tabular}[c]{@{}c@{}}Travel\\Booking\end{tabular}
& Average \\
\midrule
Gemini-3-Flash & \underline{13.44} & \underline{13.55} & 11.98 & 13.68 & 13.15 \\
GPT-5-mini     & \textbf{12.90}    & \textbf{12.48}    & 10.52 & 10.71 & \textbf{11.66} \\
DeepSeek-V3.2  & 15.90             & 13.88             & 10.78 & 10.64 & 12.80 \\
\midrule
BM25           & 17.04             & 14.52             & 10.61 & 11.18 & 13.35 \\
Qwen3-Emb-4B   & 16.28             & 14.36             & \underline{10.15} & 11.20 & 13.06 \\
GraphRAG       & 16.72             & 14.54             & 10.28 & 10.90 & 13.11 \\
\midrule
MemAgent       & 15.74             & 13.68             & 11.08 & \textbf{10.32} & \underline{12.71} \\
MemoBrain      & 16.00             & 13.96             & 10.42 & 10.60 & 12.74 \\
\midrule
Mem0           & 16.20             & 13.98             & 10.58 & 10.86 & 12.90 \\
A-MEM          & 16.06             & 14.28             & 10.65 & 11.20 & 13.06 \\
MemOS          & 16.02             & 14.04             & 10.49 & 10.80 & 12.85 \\
MemoryOS       & 16.04             & 13.88             & \textbf{9.06} & 11.68 & 13.04 \\
\midrule
AWM            & 19.88             & 15.86             & 10.64 & \underline{10.56} & 14.24 \\
SkillWeaver    & 17.14             & 14.16             & 10.94 & 10.96 & 13.30 \\
AgentKB        & 30.48             & 14.74             & 14.87 & 12.70 & 18.25 \\
ACE            & 19.00             & 14.84             & 11.67 & 11.58 & 14.29 \\
ReasoningBank  & 21.12             & 15.10             & 11.42 & 12.40 & 15.01 \\
\midrule
MemEvolve      & 16.32             & 14.64             & 11.53 & 11.44 & 13.51 \\
\bottomrule
\end{tabular}

\end{table*}

\begin{table*}[htbp]
\centering
\caption{Efficiency comparison on \textsc{InEp-Exec} under the 128K setting, measured by token usage. Lower is better. The best result in each column is bolded and the runner-up is underlined.}
\label{tab:bfcl_ie_longcontext_token_128k}
\setlength{\tabcolsep}{5pt}
\renewcommand{\arraystretch}{0.8}

\begin{tabular}{lccccc}
\toprule
Method
& \begin{tabular}[c]{@{}c@{}}Gorilla File\\System\end{tabular}
& \begin{tabular}[c]{@{}c@{}}Vehicle\\Control\end{tabular}
& \begin{tabular}[c]{@{}c@{}}Trading\\Bots\end{tabular}
& \begin{tabular}[c]{@{}c@{}}Travel\\Booking\end{tabular}
& Average \\
\midrule
Gemini-3-Flash & \underline{60136} & \underline{82076} & 336793 & 80528 & 141437 \\
GPT-5-mini     & \textbf{46875}    & \textbf{67273}    & \textbf{178668} & \textbf{50309} & \textbf{85959} \\
DeepSeek-V3.2  & 83920             & 96226             & 199725 & 64076 & \underline{110987} \\
\midrule
BM25           & 141335            & 162259            & 261983 & 105814 & 167848 \\
Qwen3-Emb-4B   & 135416            & 174900            & 279903 & 110877 & 175274 \\
GraphRAG       & 144339            & 167429            & 215260 & 106440 & 158367 \\
\midrule
MemAgent       & 84212             & 92239 & 224752 & \underline{61541} & 115686 \\
MemoBrain      & 131772            & 131863            & \underline{197039} & 81771 & 135611 \\
\midrule
Mem0           & 133638            & 155550            & 229999 & 117347 & 159133 \\
A-MEM          & 118105            & 132459            & 221201 & 106099 & 144080 \\
MemOS          & 105597            & 123065            & 239596 & 83622 & 137459 \\
MemoryOS       & 142679            & 219548            & 215285 & 158213 & 180424 \\
\midrule
AWM            & 136952            & 163160            & 224228 & 75445 & 149946 \\
SkillWeaver    & 99893             & 104069            & 218444 & 71885 & 123573 \\
AgentKB        & 227553            & 116884            & 395484 & 95346 & 205974 \\
ACE            & 157982            & 177453            & 326612 & 126807 & 196563 \\
ReasoningBank  & 156512            & 141968            & 240103 & 97710 & 189712 \\
\midrule
MemEvolve      & 95391             & 117017            & 280024 & 78307 & 140593 \\
\bottomrule
\end{tabular}

\end{table*}

\clearpage
\section*{NeurIPS Paper Checklist}

\begin{enumerate}

\item {\bf Claims}
    \item[] Question: Do the main claims made in the abstract and introduction accurately reflect the paper's contributions and scope?
    \item[] Answer: \answerYes{} %
    \item[] Justification: The abstract and introduction state the three contributions of the paper (the self-evolving perspective on memory, the four evaluation settings, and the \EMB{} benchmark), and these claims are supported by the formulation in Section~3, the benchmark construction in Section~4, and the empirical analysis in Section~5.

\item {\bf Limitations}
    \item[] Question: Does the paper discuss the limitations of the work performed by the authors?
    \item[] Answer: \answerYes{} %
    \item[] Justification: The limitations of this paper are presented in Section 6.

\item {\bf Theory assumptions and proofs}
    \item[] Question: For each theoretical result, does the paper provide the full set of assumptions and a complete (and correct) proof?
    \item[] Answer: \answerNA{} %
    \item[] Justification: The paper presents an empirical benchmark study and does not include theoretical results, theorems, or formal proofs.

    \item {\bf Experimental result reproducibility}
    \item[] Question: Does the paper fully disclose all the information needed to reproduce the main experimental results of the paper to the extent that it affects the main claims and/or conclusions of the paper (regardless of whether the code and data are provided or not)?
    \item[] Answer: \answerYes{} %
    \item[] Justification: Section~4 describes the source datasets and our reconstruction pipeline, Section~5.1 describes the LLM backbones, the categories of memory methods, the unified \texttt{utilize}/\texttt{update} interface, the per-setting update schedule, the context budgets, the embedding backend, and the retrieval top-$k$. Additional dataset reconstruction prompts are provided in the appendix.

\item {\bf Open access to data and code}
    \item[] Question: Does the paper provide open access to the data and code, with sufficient instructions to faithfully reproduce the main experimental results, as described in supplemental material?
    \item[] Answer: \answerYes{} %
    \item[] Justification: Our data and code are available at https://github.com/DSAIL-Memory/EvoMemBench.

\item {\bf Experimental setting/details}
    \item[] Question: Does the paper specify all the training and test details (e.g., data splits, hyperparameters, how they were chosen, type of optimizer) necessary to understand the results?
    \item[] Answer: \answerYes{} %
    \item[] Justification: We do not train new models. We specify the LLM backbones, dataset splits, evaluation protocols, memory update schedules for each setting, context budgets (16K–128K) for INEP-EXEC, the default embedding model (text-embedding-v4), and retrieval top-k values (10 for knowledge-oriented tasks and 3 for execution-oriented tasks).

\item {\bf Experiment statistical significance}
    \item[] Question: Does the paper report error bars suitably and correctly defined or other appropriate information about the statistical significance of the experiments?
    \item[] Answer: \answerYes{} %
    \item[] Justification: For each setting, results are aggregated over a large number of evaluation samples spanning multiple sub-datasets and sub-domains, and we additionally report average rank across sub-datasets so that the comparison reflects consistent trends rather than single-sample fluctuations. The findings highlighted in the paper correspond to gaps that are large and consistent across these multiple sub-datasets, context budgets, and difficulty levels.

\item {\bf Experiments compute resources}
    \item[] Question: For each experiment, does the paper provide sufficient information on the computer resources (type of compute workers, memory, time of execution) needed to reproduce the experiments?
    \item[] Answer: \answerYes{} %
    \item[] Justification: All evaluations are performed by calling the served LLM APIs (DeepSeek-V3.2) and the DashScope embedding API; we do not train models. Token usage, which is the dominant cost driver, is reported as an efficiency metric alongside task performance. The orchestration and memory modules run on a single CPU machine.

\item {\bf Code of ethics}
    \item[] Question: Does the research conducted in the paper conform, in every respect, with the NeurIPS Code of Ethics \url{https://neurips.cc/public/EthicsGuidelines}?
    \item[] Answer: \answerYes{} %
    \item[] Justification: The work uses publicly available datasets and API-served LLMs to construct an agent-memory benchmark, does not involve human subjects or personal data, and complies with the NeurIPS Code of Ethics.

\item {\bf Broader impacts}
    \item[] Question: Does the paper discuss both potential positive societal impacts and negative societal impacts of the work performed?
    \item[] Answer: \answerNA{} %
    \item[] Justification: \EMB{} is a measurement benchmark for agent memory and does not introduce a new model or deployable system. Any positive impact (better evaluation of memory) and negative impact (more capable autonomous agents) is indirect and shared with the broader literature on LLM agents.
    
\item {\bf Safeguards}
    \item[] Question: Does the paper describe safeguards that have been put in place for responsible release of data or models that have a high risk for misuse (e.g., pre-trained language models, image generators, or scraped datasets)?
    \item[] Answer: \answerNA{} %
    \item[] Justification: We do not release pre-trained generative models or scraped data. \EMB{} is built by reusing or reconstructing existing public benchmarks whose contents do not pose a high risk of misuse.

\item {\bf Licenses for existing assets}
    \item[] Question: Are the creators or original owners of assets (e.g., code, data, models), used in the paper, properly credited and are the license and terms of use explicitly mentioned and properly respected?
    \item[] Answer: \answerYes{} %
    \item[] Justification: All source datasets (MemoryAgentBench, BFCL, CL-Bench, xbench-DeepSearch, WebWalkerQA, ALFWorld) and all evaluated memory methods and LLM backbones are cited to their original works, and we use them under their original licenses and intended research use.

\item {\bf New assets}
    \item[] Question: Are new assets introduced in the paper well documented and is the documentation provided alongside the assets?
    \item[] Answer: \answerYes{} %
    \item[] Justification: The new asset is the \EMB{} benchmark, including the reconstructed datasets and the unified evaluation harness. Documentation covers dataset construction (Section~3 and the appendix), the four evaluation settings, the \texttt{utilize}/\texttt{update} memory interface, and the per-setting evaluation protocol; the released repository will include README files and example scripts.

\item {\bf Crowdsourcing and research with human subjects}
    \item[] Question: For crowdsourcing experiments and research with human subjects, does the paper include the full text of instructions given to participants and screenshots, if applicable, as well as details about compensation (if any)?
    \item[] Answer: \answerNA{} %
    \item[] Justification: The paper does not involve crowdsourcing or research with human subjects; dataset reconstruction is performed automatically using LLMs and verifiable rules.

\item {\bf Institutional review board (IRB) approvals or equivalent for research with human subjects}
    \item[] Question: Does the paper describe potential risks incurred by study participants, whether such risks were disclosed to the subjects, and whether Institutional Review Board (IRB) approvals (or an equivalent approval/review based on the requirements of your country or institution) were obtained?
    \item[] Answer: \answerNA{} %
    \item[] Justification: The paper does not involve human subjects, so IRB approval is not applicable.

\item {\bf Declaration of LLM usage}
    \item[] Question: Does the paper describe the usage of LLMs if it is an important, original, or non-standard component of the core methods in this research? Note that if the LLM is used only for writing, editing, or formatting purposes and does \emph{not} impact the core methodology, scientific rigor, or originality of the research, declaration is not required.
    \item[] Answer: \answerYes{} %
    \item[] Justification: LLMs are central to this work: they serve as the agent backbones being evaluated (Gemini-3-Flash, GPT-5-mini, DeepSeek-V3.2) and are also used in the dataset reconstruction pipeline for \textsc{InEp-Exec} and \textsc{CrossEp-Tool}. These usages are explicitly described in Section~4 and Section~5, with reconstruction prompts provided in the appendix.

\end{enumerate}

\end{document}